\documentclass[acmsmall]{acmart}
\makeatletter
\def\runningfoot{\def\@runningfoot{}}
\def\firstfoot{\def\@firstfoot{}}
\makeatother
\renewcommand\footnotetextcopyrightpermission[1]{} 
\usepackage{fancyhdr}
\AtBeginDocument{%
  \providecommand\BibTeX{{%
    \normalfont B\kern-0.5em{\scshape i\kern-0.25em b}\kern-0.8em\TeX}}}

\usepackage{amssymb}
\usepackage{multirow}
\usepackage{xcolor}
\usepackage{colortbl}  
\newcommand{\ie}{\textit{i}.\textit{e}.}
\newcommand{\eg}{\textit{e}.\textit{g}.}
\newcommand{\cf}{\textit{cf.}}

\usepackage{pdfrender}

\usepackage{bm}
\usepackage{amsmath}
\usepackage[caption=false,font=footnotesize,labelfont=sf,textfont=sf]{subfig}
\usepackage[switch]{lineno}
\usepackage{url}
\usepackage{graphicx}
\usepackage{booktabs}
\usepackage{arydshln}

\newcommand{\cc}[1]{{\color{black}{#1}}}

\begin{document}

\title{VL-NMS: Breaking Proposal Bottlenecks in Two-Stage Visual-Language Matching}

\author{Chenchi Zhang}
\authornote{Both authors contributed equally to this research.}
\email{chenchiz@zju.edu.cn}
\affiliation{%
  \institution{Zhejiang University}
  \city{Hangzhou}
  \state{Zhejiang}
  \country{China}
  \postcode{310027}
}

\author{Wenbo Ma}
\authornotemark[1]
\email{mwb@zju.edu.cn}
\affiliation{%
  \institution{Zhejiang University}
  \city{Hangzhou}
  \state{Zhejiang}
  \country{China}
  \postcode{310027}
}

\author{Jun Xiao}
\email{junx@zju.edu.cn}
\affiliation{%
 \institution{Zhejiang University}
  \city{Hangzhou}
  \state{Zhejiang}
  \country{China}
  \postcode{310027}
}

\author{Hanwang Zhang}
\email{hanwangzhang@ntu.edu.sg}
\affiliation{%
  \institution{Nanyang Technological University}
  \city{Singapore}
  \postcode{639798}
  \country{Singapore}
}

\author{Jian Shao}
\email{jshao@cs.zju.edu.cn}
\affiliation{%
 \institution{Zhejiang University}
  \city{Hangzhou}
  \state{Zhejiang}
  \country{China}
  \postcode{310027}
}

\author{Yueting Zhuang}
\email{yzhuang@zju.edu.cn}
\affiliation{%
 \institution{Zhejiang University}
  \city{Hangzhou}
  \state{Zhejiang}
  \country{China}
  \postcode{310027}
}

\author{Long Chen}
\authornote{Long Chen is the corresponding author.}
\email{zjuchenlong@gmail.com}
\affiliation{%
  \institution{Hong Kong University of Science and Technology}
  \city{Hong Kong}
  \country{China}
  \postcode{999077}
}

\renewcommand{\shortauthors}{Zhang and Ma, et al.}

\begin{abstract}
  The prevailing framework for matching multimodal inputs is based on a two-stage process: 1) detecting proposals with an object detector and 2) matching text queries with proposals. Existing two-stage solutions mostly focus on the matching step. In this paper, we argue that these methods overlook an obvious \emph{mismatch} between the roles of proposals in the two stages: they generate proposals solely based on the detection confidence (\ie, query-agnostic), hoping that the proposals contain all instances mentioned in the text query (\ie, query-aware). Due to this mismatch, chances are that proposals relevant to the text query are suppressed during the filtering process, which in turn bounds the matching performance. To this end, we propose VL-NMS, which is the first method to yield query-aware proposals at the first stage. VL-NMS regards all mentioned instances as critical objects, and introduces a lightweight module to predict a score for aligning each proposal with a critical object. These scores can guide the NMS operation to filter out proposals irrelevant to the text query, increasing the recall of critical objects, and resulting in a significantly improved matching performance. Since VL-NMS is agnostic to the matching step, it can be easily integrated into any state-of-the-art two-stage matching method. We validate the effectiveness of VL-NMS on three multimodal matching tasks, namely referring expression grounding, phrase grounding, and image-text matching. Extensive ablation studies on several baselines and benchmarks consistently demonstrate the superiority of VL-NMS.
\end{abstract}



\maketitle

\section{Introduction}
Recently, visual-language tasks which require joint reasoning over both vision and natural language modalities have attracted great interest in the multimedia community. Matching multimodal inputs is one of the most fundamental abilities that can facilitate many high-level visual-language tasks. Considering different granularities of matching, there are two types of tasks: 1) \textbf{Instance-level matching}, where each text query refers to one targeted instance (referent) in the image and is explicitly grounded to one image region, \eg, referring expression grounding (REG)~\cite{yu2018mattnet,liu2019learning,liu2019improving} or phrase grounding (PG)~\cite{yu2018rethinking}. 2) \textbf{Image-level matching}, where text queries describe the whole image and are matched with images holistically, \eg, image-text matching (ITM)~\cite{lee2018stacked,chen2020imram,liu2020graph}. Both two types of visual-language matching tasks are important for many downstream applications such as bidirectional cross-modal retrieval~\cite{feng2014cross}, visual question answering~\cite{antol2015vqa}, visual navigation~\cite{chen2019touchdown} and autonomous driving~\cite{kim2019grounding}.

\begin{figure}[t]
	\centering
	\includegraphics[width=1\linewidth]{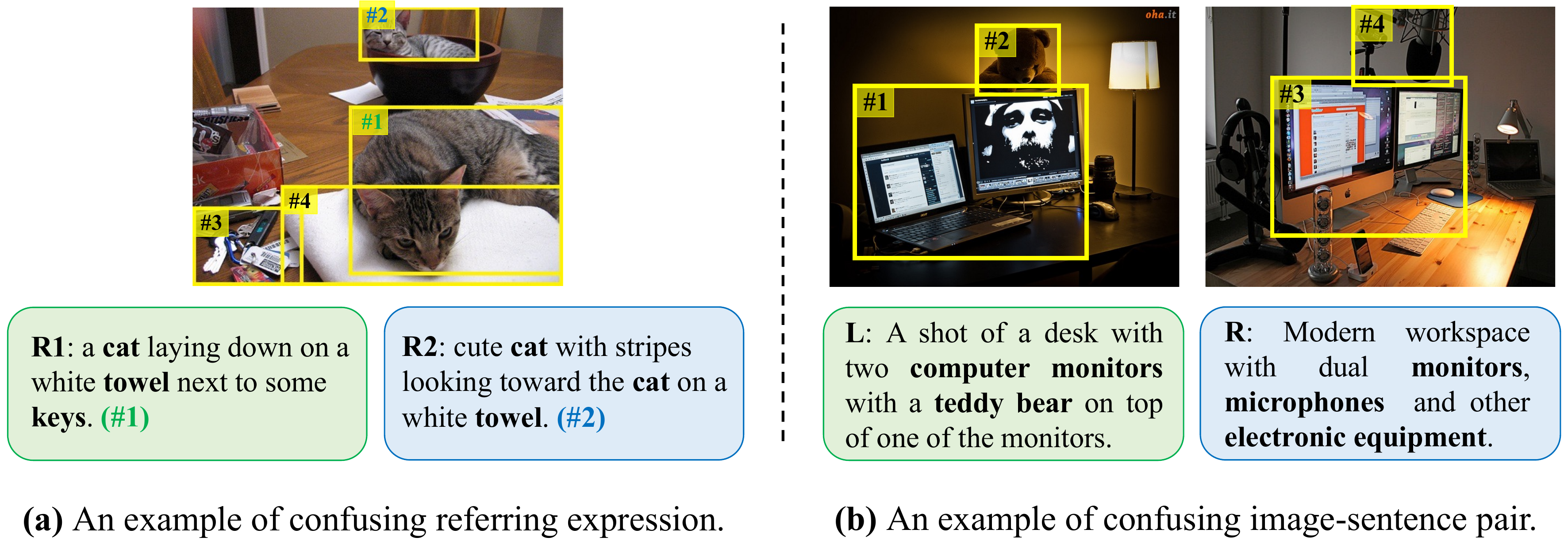}
	\caption{Confusing examples from REG and ITM tasks. (a): A referring expression comprehension example from RefCOCOg~\cite{mao2016generation}. Two similar expressions (R1 and R2) refer to different objects. (b): An image-text matching example from MSCOCO~\cite{lin2014microsoft}. Two similar images with overlapping visual content.}
	\label{fig:motivation_a}
 \vspace{-15pt}
\end{figure}

The vast majority of state-of-the-art visual-language matching methods are in a two-stage manner. In the first stage, a pretrained object detector is used to extract region proposals from images. At the second stage, these proposals are either treated as candidate bounding boxes (bboxes) to be selected from as in instance-level matching (\eg, REG or PG), or regraded as a bottom-up attention mechanism~\cite{anderson2018bottom} as in image-level matching (\eg, ITM).

Specifically, two-stage instance-level matching methods normally divide the input query into components, ground each component to one proposal and reason about the interactions among them under the guidance of the input query. Take the REG as an example, compared with one-stage grounding methods which regard REG as a generalized object detection (or segmentation) task, two-stage REG methods with the ``detect-and-ground'' pipeline is more similar to the human way of reasoning. More importantly, it's much easier for two-stage methods to exploit the linguistic structure of referring expressions and perform global reasoning, which is vital when dealing with long, complex expressions. For example in Fig.~\ref{fig:motivation_a}(a), when grounding ``\emph{a cat laying down on a white towel next to some keys}", it is even difficult for humans to identify the referent \texttt{cat} without considering its contextual objects \texttt{towel} and \texttt{keys}. Also, it has been pointed out that one-stage methods are insensitive to linguistic variations. When changing the expression in Fig.~\ref{fig:motivation_a}(a) to ``\emph{cute cat with stripes looking toward the cat on a white towel}", they tend to refer to the same object (\#1)~\cite{akula2020words}. In general, two-stage methods with perfect proposals (\eg, all human-annotated object regions) can achieve more accurate and explainable grounding results than their one-stage counterparts.

On the other hand, two-stage image-level matching methods cross-match region proposals with query words and use these local matching scores to measure the global image-sentence similarity. For example, it has been widely recognized that by inferring the latent correspondence between each image region and each query word, two-stage ITM methods can better capture the fine-grained interplay between vision and language and are more interpretable to human~\cite{lee2018stacked}. As shown in Fig.~\ref{fig:motivation_a}(b), these two images contain overlapping visual content (\texttt{monitors}) which makes it difficult to distinguish them from one another on the image level. But by paying attention to other critical objects (\eg, \texttt{teddy bear} in the left image, \texttt{microphones} in the right image), two-stage methods have a better chance to match them with the right sentence.

\begin{figure}[t]
	\centering
	\includegraphics[width=0.43\linewidth]{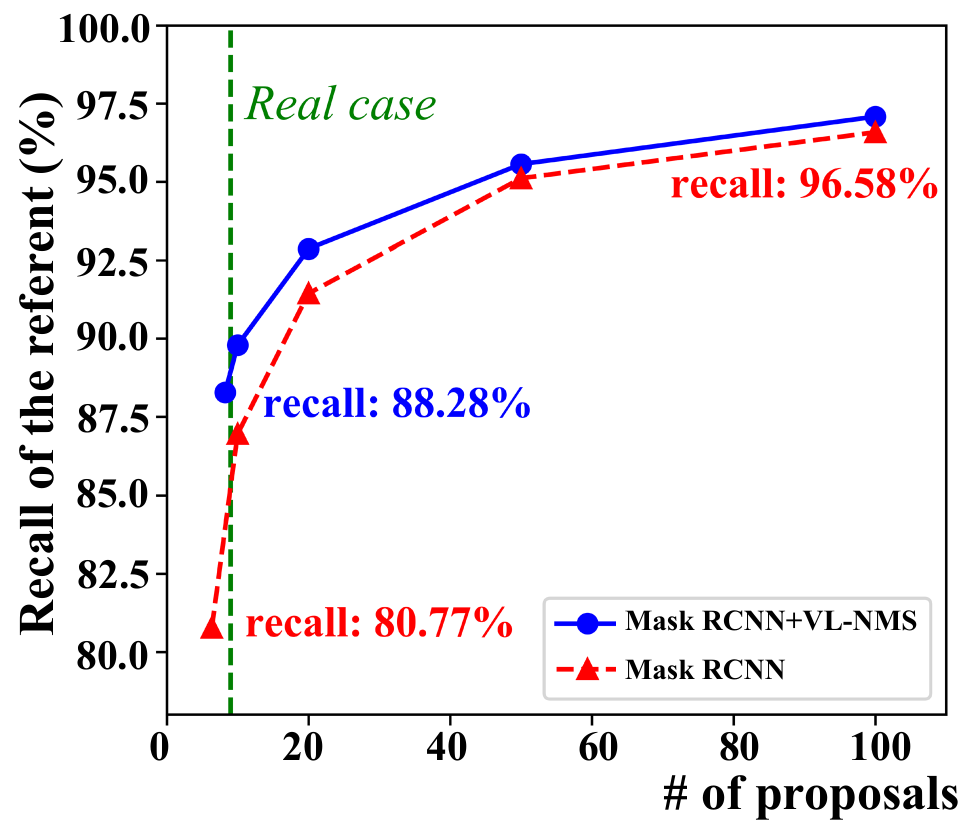}
	\caption{The recall of the referent (IoU$>$0.5) vs. number of proposals on the RefCOCO testB set. The real case denotes the actual situation in all state-of-the-art two-stage grounding methods.}
	\label{fig:motivation_b}
 \vspace{-15pt}
\end{figure}

\begin{figure}[t]
	\centering
	\includegraphics[width=0.4\linewidth]{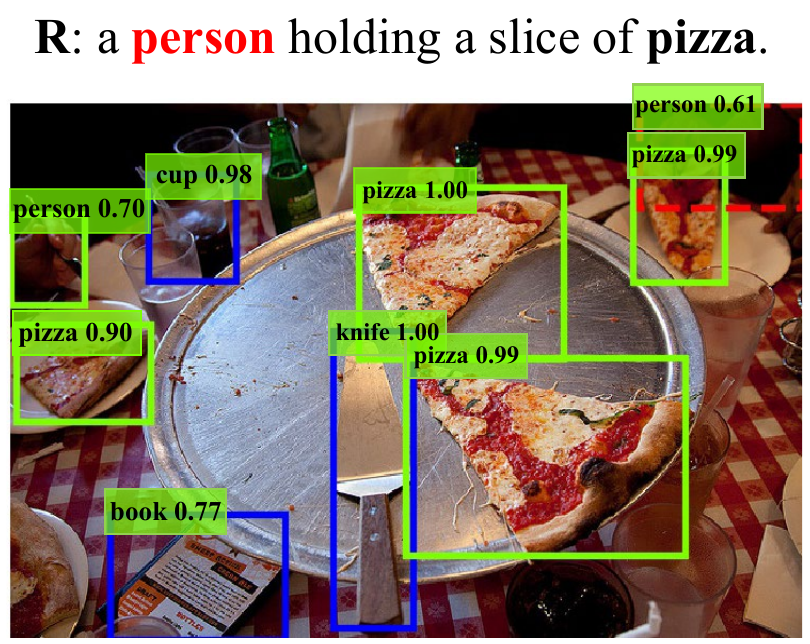}
	\caption{An example of the first-stage proposals of prevailing MAttNet~\cite{yu2018mattnet}. The proposals only contain bboxes with high detection confidence regardless of the content of expression (\eg, The candidates \texttt{knife}, \texttt{book}, and \texttt{cup} are not mentioned in expression). The bbox in red dashed line denotes the missing referent.}
	\label{fig:motivation_c}
 \vspace{-15pt}
\end{figure}

However, the performance of two-stage visual-language matching methods is heavily bounded by the proposal quality. Especially for REG, when changing the proposals from human-annotated regions to detected regions of a pretrained detector, two-stage REG methods' performance drops dramatically. In this paper, we argue that this huge performance gap between the detected and ground-truth proposals is mainly caused by the \textbf{mismatch} between the roles of proposals in the two stages: \emph{the first-stage network generates proposals solely based on detection confidence, while the second-stage network just assumes that the generated proposals will contain all instances mentioned in the text query}. Take REG as an example. For each image, a well pre-trained detector can detect hundreds of detections with a near-perfect recall of the referent and contextual objects (\eg, as shown in Fig.~\ref{fig:motivation_b}, recall of the referent can reach up to 96.58\% with top-100 detections). However, to relieve the burden of the referent grounding step in the second stage, current two-stage methods always filter proposals simply based on their detection confidences. These heuristic rules result in a sharp reduction of the recall (\eg, decrease to 80.77\% as in Fig.~\ref{fig:motivation_b}), and bring in the mismatch negligently. To illustrate this further, we show a concrete example in Fig.~\ref{fig:motivation_c}. To ground the referent at the second stage, we hope that the proposals contain the referent \texttt{person} and its contextual object \texttt{pizza}. In contrast, the first-stage network only keeps bboxes with high detection confidence (\eg, \texttt{knife}, \texttt{book}, and \texttt{cup}) as proposals, but misses the critical referent \texttt{person} (\ie, the red bbox).

In this paper, we propose a novel algorithm VL-NMS, to rectify the mismatch of detected proposals at the conventional first stage. In particular, for each text query, VL-NMS regards all nouns in the text query as critical objects, and introduces a lightweight relatedness module to predict a probability score for each proposal to be a critical object. The higher predicted score denotes the higher relevance between a proposal and the text query. Then, we fuse the relatedness scores and classification scores, and adopt the fused scores as the suppression criterion in Non-Maximum Suppression (NMS). After NMS, we can filter out the proposals with little relevance to the text query. Finally, all proposals and the text query are fed into the second-stage grounding or matching network, to obtain the final prediction.

Extensive ablations illustrate the superiority of VL-NMS on these visual-language matching tasks. We demonstrate through experiments that VL-NMS can learn the transferable fine-grained correspondence between visual and language. This characteristic is extremely useful when bbox annotations are absent from the target tasks (\eg, the image-text matching tasks). The contributions and novelty of this work are summarized as follows:
\begin{itemize}
\item We are the first to point out the proposal bottlenecks in two-stage visual-language matching methods. By investigating the recall of critical objects, we attribute the degraded performance of two-stage methods to the query-agnostic proposals generated at the first stage.
\item We propose a novel algorithm VL-NMS to rectify the mismatch of proposals at the first stage by making the proposals query-aware. We also explore different ways to build pseudo ground-truth for training VL-NMS.
\item We demonstrate significant performance gains on three REG benchmarks, one phrase grounding benchmark, and one ITM benchmark. It's worth noting that VL-NMS can be generalized and easily integrated into any state-of-the-art two-stage grounding or matching methods to further boost their performance.
\item Our method is efficient, generalizable, and transferable, opening doors for many downstream applications such as multimodal summarization.
\end{itemize}

\noindent\textbf{Highlight.} It is worth noting that this paper is a substantial extension of our previous AAAI conference publication~\cite{chen2021ref}. Compared to the original conference version, this new manuscript has made three main improvements: 1) We explore an alternative way in the relatedness module design based on the multi-modal BERT (dubbed VL-NMS$_{\text{Trans}}$). The results show that the new method further surpasses the original performance. 2) We propose a new approach to build pseudo ground-truth by means of weakly supervised phrase grounding. Training VL-NMS no longer needs bbox annotations from the COCO-detection dataset. 3) We extend VL-NMS to other instance-level matching task phrase grounding and image-level matching task image-text matching. Extensive ablations illustrate the superiority of VL-NMS on different visual-language matching tasks.

\begin{figure*}[t]
	\centering
	\includegraphics[width=1\linewidth]{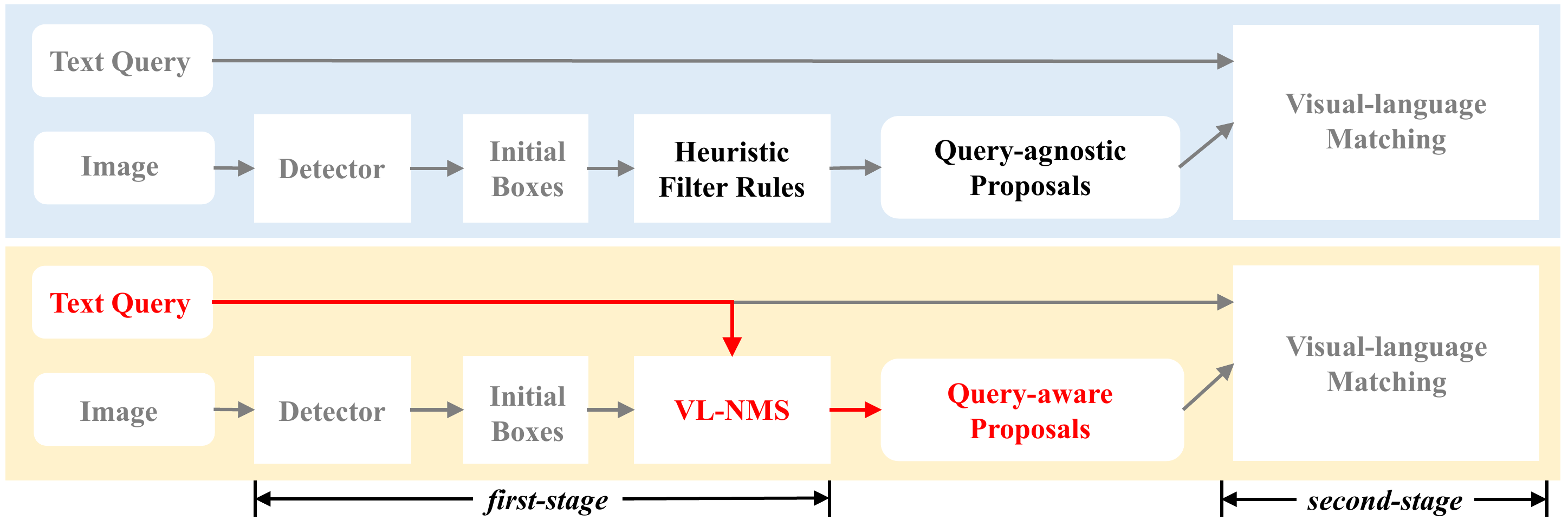}
	\caption{Upper: A typical two-stage visual-language matching framework, which uses heuristic filter rules to obtain query-agnostic proposals at the first-stage. Below: The VL-NMS module can generate query-aware proposals by considering the text query at the first stage.}
	\label{fig:two-stage-models}
 \vspace{-15pt}
\end{figure*}

\section{Related Work}

\subsection{Referring Expression Grounding}
Considering different granularities of localization, there are two types of REG: 1) Referring Expression Comprehension (REC)~\cite{hu2017modeling,yu2016modeling,yu2017joint,8307406}, where the referents are localized by bboxes. 2) Referring Expression Segmentation (RES)~\cite{hu2016segmentation,shi2018key,8845685,8978485}, where the referents are localized by masks.
\subsubsection{Referring Expression Comprehension}
The current overwhelming majority of REC methods are in a two-stage manner: proposal generation and referent grounding. To the best of our knowledge, existing two-stage works all focus on the second stage. Specifically, they tend to design a more explainable reasoning process by structural modeling~\cite{yu2018mattnet,liu2019improving,liu2019learning,liu2019joint,hong2019learning,niu2019variational}, or more effective multi-modal interaction mechanism~\cite{wang2019neighbourhood,yang2020graph}. However, their performance is strictly limited by the proposals from the first stage. Recently, another emerging direction to solve REC is in a one-stage manner~\cite{chen2018real,yang2019fast,liao2020real,luo2020multi,yang2020improving}. Although one-stage methods achieve faster inference speed empirically, they come at a cost of lost interpretability and poor performance in composite expressions. \cc{Compared with the one-stage approaches, we both try to achieve text and image communication at an early stage. The difference is that one-stage directly outputs the only referent target, while VL-NMS outputs all objects mentioned in the expression. These query-aware proposals from VL-NMS are closely related to the query (\ie, expression), which can significantly improve both the accuracy and explainability of the second-stage model reasoning. More recently, Transformer has been applied in visual grounding~\cite{kamath2021mdetr, deng2021transvg,zhu2022seqtr, yang2022improving} with excellent performance. But there is still the problem of poor interpretability. Besides, the vast majority of methods require a lot of prior knowledge and expertise to design sophisticated networks. This makes them less generalizable. In addition to the methods above, some recently published works~\cite{9859830,ye2022shifting} are inspired by our previous conference paper. QRNet~\cite{ye2022shifting} shares a similar idea with us and tries to resolve the inconsistency in features. But we still have differences, most notably we are based on the two-stage model in order to achieve better interpretability. In this paper, we rectify the overlooked mismatch in two-stage methods and propose VL-NMS. It is worth noting that VL-NMS can be attached on any two-stage model easily and the effect is validated on multiple models and benchmarks.}
\subsubsection{Referring Expression Segmentation} 
Unlike REC, most RES works are one-stage methods. They typically utilize a ``concatenation-convolution'' design to combine the two different modalities: they first concatenate the expression feature with visual features at each location, and then use several conv-layers to fuse the multimodal features for mask generation. To further improve mask qualities, they usually enhance their backbones with more effective features by multi-scale feature fusion~\cite{margffoy2018dynamic}, feature progressive refinement~\cite{li2018referring,chen2019see,huang2020referring}, or novel attention mechanisms~\cite{shi2018key,ye2019cross,hu2020bi}. Besides, with the development of two-stage instance segmentation (\eg, Mask R-CNN~\cite{he2017mask}), two-stage REC methods can be extended to solve RES simply by applying the mask branch upon the predicted bounding box at the end of the second stage. Analogously, VL-NMS can be easily integrated into any two-stage RES method.

\subsection{Image-Text Matching} 
Image-Text matching aims to align images with sentences, normally evaluated by the recall of bidirectional retrieval. Prior works directly learn the global correspondence between images and sentences~\cite{8777191,faghri2017vse++,wu2018learning,xu2021cross}. These works normally encode images and sentences into holistic feature vectors, project them into a common feature space and try to maximize the similarity between the matched image-sentence pairs. Recent works propose to infer the latent local correspondence between each image region and each word or noun chunk in the sentence~\cite{lee2018stacked,chen2020imram,liu2020graph}. By exploiting this fine-grained correspondence, the global matching performance can be improved. This type of method typically uses a pretrained object detector to extract region proposals from images, then match them with each word in the sentence. VL-NMS can help to filter out region proposals irrelevant to the sentence and aid the local correspondence inference process.

\subsection{Phrase Grounding} 
Given an image and a sentence, phrase grounding aims to ground all noun phrases in the sentence. There are also two types of solutions: proposal-free methods and proposal-driven methods. Different from REC, the queries in phrase grounding are much simpler, which relieves two-stage methods from complicated relational reasoning and allows them to take more proposals at the second stage\footnote{At the second stage, REC methods take an average number of 10 proposals while phrase grounding methods can take more than 200 proposals.}. Meanwhile, efforts have been taken to handle the query diversity at the first stage by either using an object detector pre-trained on another large-scale dataset~\cite{yu2018rethinking} or re-generate proposals with respect to queries and mentioned objects~\cite{chen2017query}. Recently, a large proportion of phrase grounding works turn to the weakly supervised setting where only image-sentence alignment is provided as annotation~\cite{chen2018knowledge,datta2019align2ground}. In this paper, we use weakly supervised phrase grounding as an alternative way to capture the cross-modal context in generated region proposals. Besides, VL-NMS is model-agnostic and thus can help various two-stage methods get better performance on phrase grounding.

\subsection{Non-Maximum Suppression} 
Non-Maximum Suppression (NMS) is a de facto standard post-processing step adopted by numerous modern object detectors, which removes duplicate bboxes based on detection confidence. Except for the most prevalent GreedyNMS, multiple improved variants have been proposed recently. Generally, they can be categorized into three groups: 1)~Criterion-based~\cite{jiang2018acquisition,tychsen2018improving,tan2019learning,yang2019learning}: they utilize other scores instead of classification confidence as the criterion to remove bboxes by NMS, \eg, IoU scores. 2)~Learning-based~\cite{hosang2017learning,hu2018relation}: they directly learn an extra network to remove duplicate bboxes. 3)~Heuristic-based~\cite{bodla2017soft,liu2019adaptive}: they dynamically adjust the thresholds for suppression according to some heuristic rules. In this paper, we are inspired by the criterion-based NMS, and design the VL-NMS, which uses both expression relatedness and detection confidence as the criterion.

\section{Approach}

In this section, we first revisit the typical two-stage visual-language matching framework, and then introduce the details about VL-NMS, including the relatedness module, the ground-truth acquisition, and the training objectives.

\subsection{Revisiting Two-Stage Visual-language Matching}
The two-stage framework is the most prevalent pipeline for visual-language matching tasks. As shown in Fig.~\ref{fig:two-stage-models}, it consists of two separate stages: proposal generation at the first stage and multimodal matching at the second stage.

\subsubsection{Proposal Generation} Given an image, current two-stage methods always resort to a well pre-trained detector to obtain a set of initially detected bboxes, and utilize an NMS operation to remove duplicate bboxes. However, even after the NMS operation, there are still thousands of bboxes left (\eg, for the REG task, each image in RefCOCO has an average of 3,500 detections). To relieve the burden of the following multimodal matching step, all existing works further filter these bboxes based on their detection confidences. Although this heuristic filter rule can reduce the number of proposals, it also results in a drastic drop in the recall of the mentioned instances. Especially for the REG task, this recall drop can be fatal as the grounding process at the second stage will be guaranteed to fail if the referent is missed at the first stage. Detailed recall statistics of the referent and contextual objects on three REG benchmarks are reported in Table~\ref{tab:recall}.

\subsubsection{Multimodal Matching} For the REG task, in the training phase, two-stage methods usually use the ground-truth regions in COCO as proposals, and the number is quite small (\eg, each image in RefCOCO has an average of 9.84 ground-truth regions). For explainable grounding, state-of-the-art two-stage grounding methods always compose these proposals into graph~\cite{yang2019dynamic,wang2019neighbourhood} or tree~\cite{liu2019learning,hong2019learning} structures. As the number of proposals increases linearly, the number of computations increases exponentially. Therefore, in the test phase, it is a must for them to filter detections at the end of the first stage.
For the ITM task, in both training and test phases, two-stage methods~\cite{lee2018stacked,chen2020imram,liu2020graph} normally utilize the bottom-up attention mechanism proposed in~\cite{anderson2018bottom} and take a fixed number of proposals as input at the second stage (normally 36 bboxes per image). Although this number is relatively large compared to REG, reducing the number of proposals can help improve the efficiency of the following image-sentence matching process and greatly shorten the retrieval time.

\begin{figure*}[t]
	\centering
	\includegraphics[width=1.0\linewidth]{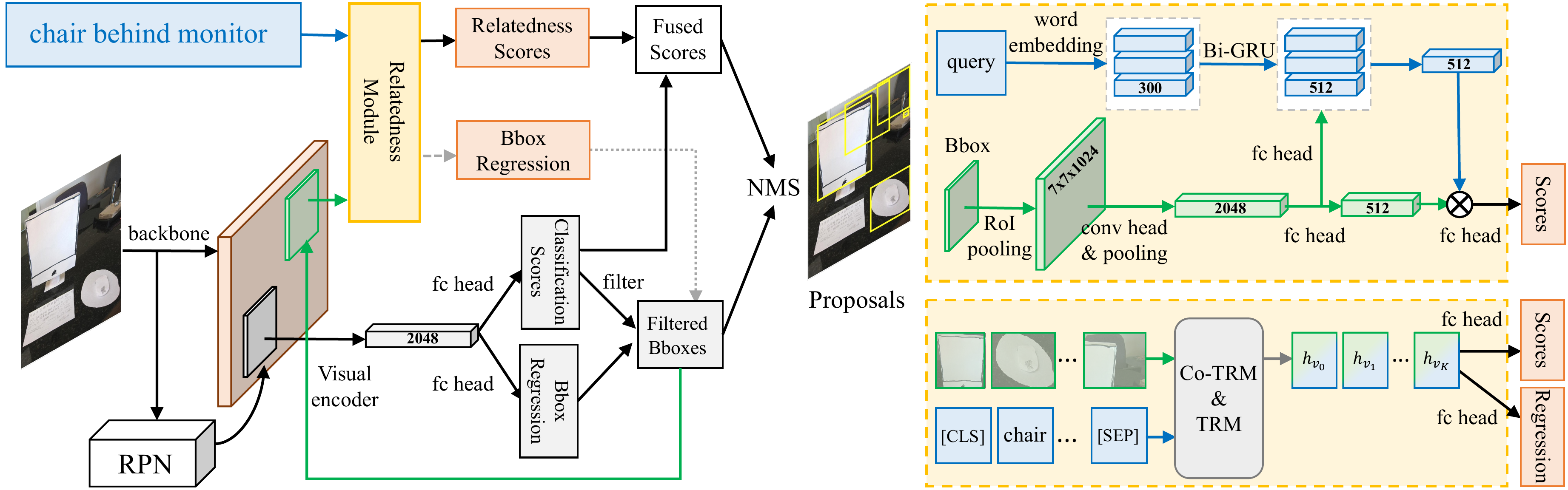}
	\caption{\textbf{Left}: The overview of the VL-NMS model. Given an image, the model uses a pre-trained detector to generate thousands of initial bboxes. Then, the text query and hundreds of filtered bboxes with detection confidence beyond a threshold are fed into the relatedness module to predict the relatedness scores. Lastly, the relatedness scores are fused with detection confidences and the fused scores are used as the suppression criterion of NMS. \textbf{Right}: Two types of relatedness module corresponding to VL-NMS$_{\text{FC}}$(top)~\cite{chen2021ref} and VL-NMS$_{\text{Trans}}$(bottom), respectively. We further introduce the regression module in VL-NMS$_{\text{Trans}}$ to improve the quality of bboxes.} 
	\label{fig:architecture}
 \vspace{-17pt}
\end{figure*}

\subsection{Relatedness Module}

An overview of the VL-NMS model is shown in Fig.~\ref{fig:architecture}. The core of VL-NMS is the relatedness module. We proposed two variants of VL-NMS: a fully-connected layer based VL-NMS$_{\text{FC}}$ and a Transformer-based VL-NMS (VL-NMS$_{\text{Trans}}$). We illustrate the two variants of VL-NMS in Fig.~\ref{fig:architecture}.

\subsubsection{VL-NMS$_{\text{FC}}$}
Given an image, a pre-trained detector can generate thousands of initial bboxes. To reduce the computation of the relatedness module, we first use a threshold $\delta$ to filter the bboxes with classification confidence, and obtain a filtered bbox set $\mathcal{B}$. For each bbox $\bm{b}_i \in \mathcal{B}$, we use a region visual encoder $e_v$ (\ie, an RoI Pooling layer and a convolutional head network) to extract the bbox feature $\bm{v}_i \in \mathbb{R}^v$. Meanwhile, for the text query $Q$, we use an text encoder $e_q$ (\ie, a Bi-GRU or BERT) to output a set of word features $\{\bm{w}_1, ..., \bm{w}_{|Q|}\}$, where $\bm{w}_j \in \mathbb{R}^q$ is the $j$-th word feature. For each bbox $\bm{b}_i$, we use a soft-attention mechanism~\cite{chen2017sca} to calculate a unique query feature $\bm{q}_i$ by:
\begin{equation} \label{eq:1}
\begin{aligned} 
\bm{v}^a_i = \text{MLP}_a (\bm{v}_i), \quad & a_{ij} = \text{FC}_s ([\bm{v}^a_i; \bm{w}_j]), \\
\alpha_{ij} = \text{softmax}_j(a_{ij}),  \quad & \bm{q}_i = \textstyle{\sum_j} \alpha_{ij} \bm{w}_j,
\end{aligned}
\end{equation}
where $\text{MLP}_a$ is a two-layer MLP mapping $\bm{v}_i \in \mathbb{R}^v$ to $\bm{v}^a_i \in \mathbb{R}^q$, $\text{FC}_s$ is a FC layer to calculate the similarity between bbox feature $\bm{v}^a_i$ and word feature $\bm{w}_j$, and $[;]$ is a concatenation operation. Then, we combine the features from both modalities and predict the relatedness score $r_i$:
\begin{equation}
\begin{aligned}
\bm{v}^b_i = \text{MLP}_b(\bm{v}_i), \quad & \bm{m}_i = \text{L2Norm}(\bm{v}^b_i \odot \bm{q}_i), \\
\hat{r}_i = \text{FC}_r(\bm{m}_i), \quad & r_i = \text{sigmoid}(\hat{r}_i),
\end{aligned}
\end{equation}
where $\text{MLP}_b$ is a two-layer MLP mapping $\bm{v}_i \in \mathbb{R}^v$ to $\bm{v}^b_i \in \mathbb{R}^q$, $\odot$ is element-wise multiplication, L2Norm is $l_2$ normalization, and $\text{FC}_r$ is a FC layer mapping $\bm{m}_i \in \mathbb{R}^q$ to $\hat{r}_i \in \mathbb{R}$.

\subsubsection{VL-NMS$_{\text{Trans}}$}
As show in Fig.~\ref{fig:architecture}, VL-NMS$_{\text{Trans}}$ is built on top of the multi-modal BERT~\cite{lu2019vilbert}, which consists of a series of Transformer blocks and co-attention layers. Considering computation and inference time, we implement a shallow, lightweight but efficient version. For the given image, we first get the bboxes feature in the same way as VL-NMS$_{\text{FC}}$. We also encode spatial locations into a 5-d vector (normalized top-left and bottom-right coordinates, and the proportion of the region to total image area). After being projected to the same dimension, the spatial location and visual feature are summed to get the representation of visual tokens. For text query $Q$, we first convert it to tokens by the tokenization step~\cite{devlin2018bert}, then put special $[\mathrm{CLS}]$ and $[\mathrm{SEP}]$ tokens at the beginning and end of the word tokens, respectively. The final text representation is the sum of a token-specific learned embedding~\cite{wu2016google}, position embedding and segment embedding. Therefore, given the image and text query, we can get the visual tokens $\{\bm{\tilde{v}}_1,\dots, \bm{\tilde{v}}_K\}$ and textual tokens $\{[\mathrm{CLS}],\bm{\tilde{w}}_1,\dots, \bm{\tilde{w}}_{|Q|},[\mathrm{SEP}]\}$ through visual and text encoder respectively, where $K$ is the number of viusal tokens. The input token sequence is $\{\bm{\tilde{v}}_1,\dots,\bm{\tilde{v}}_K,[\mathrm{CLS}],\bm{\tilde{w}}_1,\dots,\bm{\tilde{w}}_{|Q|},[\mathrm{SEP}] \}$, which can be obtained by concatenating visual tokens and textual tokens. The multi-modal feature then could be fused via a two-layer Transformer and co-attention network, and the output sequences for visual and textual tokens are $\{\bm{h}_{\tilde{v}_1}, \dots, \bm{h}_{\tilde{v}_K}\}$ and $\{ \bm{h}_{\tilde{w}_1},\dots,\bm{h}_{\tilde{w}_{|Q|}} \}$, respectively. To refine the bbox, we further attach a regression module to each visual token $\bm{h}_{\tilde{v}_i}$. 
\begin{equation}
\begin{aligned}
r_i = \text{sigmoid}(\text{FC}(h_{\tilde{v}_i})), \quad & t_i = \text{MLP}_t(h_{\tilde{v}_i}),
\end{aligned}
\end{equation}
where $\text{MLP}_t$ is composed of two ReLU layers and a linear layer. FC layer maps the $h_{\hat{v}_i}$ to a scalar. $r_i$  and $t_i=(t^x_i,t^y_i,t^w_i,t^h_i)$ denote the prediction of relatedness score and bbox offsets respectively.

After obtaining the relatedness score $r_i$ for bbox $\bm{b}_i$, we multiply $r_i$ with the classification confidence $c_i$ for bbox $\bm{b}_i$ from the original detector, and utilize the product of two scores $s_i$ as the suppression criterion of the NMS operation, \ie, $s_i = r_i \times c_i$. We argue that fusing by multiplication is straightforward, doesn't introduce any extra hyper-parameters, and yields competing empirical results over other methods, such as weighted summarization.

\noindent\textbf{Highlight.} Inspired by the success of Transformer~\cite{vaswani2017attention} in different vision tasks, we introduce an advanced VL-NMS$_{\text{Trans}}$ based on the Transformer in this new manuscript. The self-attention mechanism in VL-NMS$_{\text{Trans}}$ inherently helps to find the latent correspondences between visual regions and words. Further, we utilize a regression module to improve the lack of accuracy of bboxes. The more precise bboxes help the two-stage grounding methods achieve better performance.

\subsection{Acquisition of Ground-Truth Annotations}

To learn the relatedness score for each bbox, we need the ground-truth annotations for all instances mentioned in the text query. However, except for the PG datasets, both REG and ITM datasets don't contain such dense instance-level annotations, making it necessary to build pseudo ground-truth.

For the REG task, current datasets only have annotations about the referent. Thus, we need to generate pseudo ground-truths for contextual objects. Specifically, we explore two alternatives:
\begin{itemize}
	\item \textbf{Text Similarity based Method.} We first assign POS tags to each word using the spaCy POS tagger and extract all nouns in the expression. Then, we calculate the cosine similarity between GloVe embeddings of extracted nouns and category names of ground-truth regions in COCO\footnote{Two-stage methods always use an object detector pretrained on COCO detection dataset. Thus, we don't use extra or more annotations.}. Lastly, we use threshold $\gamma$ to filter regions as the pseudo ground-truths. 
	\item \textbf{WSPG based Method.} To relieve two-stage REG methods from depending on COCO detection annotations, we propose to use a weakly supervised phrase grounding (WSPG) method to generate pseudo ground-truth for contextual objects.
	Note that the performance of WSPG methods is far from ideal. Hence directly using the results of a WSPG model as the first stage proposals is not feasible.
\end{itemize} 
In the training phase, we regard all the pseudo ground-truth bboxes and annotated referent bboxes as foreground bboxes.

For the ITM task, we can use annotations from phrase grounding datasets as our ``golden'' pseudo ground-truth. Alternatively, to relieve VL-NMS from using extra annotations from other tasks, we propose to transfer the VL-NMS model pretrained on the REG dataset to the ITM task at inference time. Detailed measures and empirical results are described in the experiment section.

\noindent\textbf{Highlight.} The three prevalent REG datasets (RefCOCO/RefCOCO+/RefCOCOg)~\cite{yu2016modeling,mao2016generation} are all build upon the COCO dataset. Therefore, either explicitly or implicitly, most state-of-the-art grounding methods make use of extra COCO annotations. To get rid of the reliance on extra COCO annotations, we propose an alternative way to build pseudo ground-truth. With the WSPG based method, extra COCO annotations are no longer necessary for training VL-NMS.

\subsection{Training Objectives for VL-NMS}

We explored two types of classification objectives and one regression objective.
\subsubsection{Classification objective} 

\textbf{1) Binary XE Loss.} For each bbox $\bm{b}_i \in \mathcal{B}$, if it has a high overlap (\ie, IoU$>$0.5) with any foreground bbox, its ground-truth relatedness score $r^*$ is set to 1, otherwise $r^*=0$. Then the relatedness score prediction becomes a binary classification problem. We can use the binary cross-entropy (XE) loss as the classification objective:
\begin{equation}
\begin{aligned}
L_{\text{cls}} = -\frac{1}{|\mathcal{B}|} \textstyle{\sum^{|\mathcal{B}|}_{i=1}} r^*_i \log(r_i) + (1-r^*_i) \log(1-r_i).
\end{aligned}
\end{equation}

\textbf{2) Ranking Loss.} Generally, if a bbox has a higher IoU with foreground bboxes, the relatedness between the bbox and expression should be higher, \ie, we can use the ranking loss as the training objectives: 
\begin{equation}
L_{\text{cls}} = \frac{1}{N} \textstyle{\sum_{(\bm{b}_i, \bm{b}_j), \rho_i < \rho_j}} \max (0, r_i - r_j + \alpha),
\end{equation}
where $\rho_i$ denotes the largest IoU value between bbox $\bm{b}_i$ and foreground bboxes, $N$ is the total number of pos-neg training pairs, and $\alpha$ is a constant to control the ranking margin, set as 0.1. To select the pos-neg pair $(\bm{b}_i, \bm{b}_j)$, we follow the sampling-after-splitting strategy~\cite{tan2019learning}. Specifically, we first divide the bbox set $\mathcal{B}$ into 6 subsets based on a quantization $q$-value: $q_i = \lceil \text{max}(0, \rho_i - 0.5)/0.1  \rceil$, \ie, bboxes with higher IoU values will have larger $q$-values. Then, all bboxes with $\rho > 0.5$ are selected as positive samples. For each positive sample, we rank all bboxes with smaller $q$-value based on the predicted relatedness scores and select the top-$h$ bboxes as negative samples.

We empirically find that training with binary XE loss is more robust while training with ranking loss has the advantage of explicitly enforcing ranks within proposals and sometimes yields better results (More results are reported in Sec.~\ref{sec:4.1}).

\subsubsection{Regression Objective}
\textbf{1) Regression Loss.} In addition to the relatedness score classification loss, we also propose a new regression loss to further improve the proposal quality at the first stage. Specifically, for the bbox regression offsets $t=(t^x,t^y,t^w,t^h)$. We adopt the parameterization~\cite{girshick2014rich} for $t$ and obtain the scale-invariant ground-truth $\hat{t}=(\hat{t^x}, \hat{t^y}, \hat{t^w}, \hat{t^h})$. Then we train the regression block with the widely used smooth L1 loss:
\begin{equation}
\begin{aligned}
L_{\text{reg}} = L_{\text{smooth-L1}}(t_i, \hat{t}_i)
\end{aligned}
\end{equation}
where $L_{\text{smooth-L1}}$ is the smooth L1 loss. 

\subsubsection{Overall Training Objective}
For VL-NMS$_{\text{FC}}$, either binary XE Loss or ranking Loss are served as the training objective. Since the model with binary XE loss works better, VL-NMS$_{\text{Trans}}$ uses binary XE loss as classification loss. Then, we can train VL-NMS$_{\text{Trans}}$ with the multi-task objective:
\begin{equation}
\begin{aligned}
L = \frac{1}{|\mathcal{B}|} {\textstyle{\sum^{|\mathcal{B}|}_{i=1}}} L_{\text{BXE}}(r_i, r^*_i) + \lambda \frac{1}{N} \textstyle{\sum^{|\mathcal{B}|}_{i=1}} r^*_i L_{\text{reg}}(t_i, \hat{t}_i)
\end{aligned}
\end{equation}
where $L_{\text{BXE}}$ and $L_{\text{reg}}$ denote binary cross-entropy loss and smooth L1 loss respectively. $\lambda$ is a coefficient of $L_{reg}$ to balance these two losses. The $L_{reg}$ is activated only for positive bboxes(\ie, $r^*_i=1$) and $N$ is the total number of positive bboxes. Although we view the pseudo ground-truths generated by WSPG based Method as foreground bbox, they will not be used to calculate regression loss for the lack of reliable annotations.

\noindent\textbf{Highlight.} The previous VL-NMS$_{\text{FC}}$ only uses classification loss to select the objects contained in the query. However, insufficiently precise bboxes can also hinder the subsequent inference steps. To this end, we expect the regression module to enhance the quality of bboxes. Thus, we introduce the smooth L1 loss to train the regression module.

\section{Experiments}

\subsection{Results on Referring Expression Grounding} \label{sec:4.1}

\textbf{Datasets.} We evaluated VL-NMS on three challenging REG benchmarks: 1) \textbf{RefCOCO}~\cite{yu2016modeling}:  It consists of 142,210 referring expressions for 50,000 objects in 19,994 images. These expressions are collected in an interactive game interface~\cite{kazemzadeh2014referitgame}, and the average length of each expression is 3.5 words. All expression-referent pairs are split into the train, val, testA, and testB sets. The testA set contains images with multiple people and the testB set contains images with multiple objects.
2) \textbf{RefCOCO+}~\cite{yu2016modeling}: It consists of 141,564 referring expressions for 49,856 objects in 19,992 images. Similar to RefCOCO, these expressions are collected from the same game interface, and have the train, val, testA, and testB splits. But different from RefCOCO, these expressions don't include the absolute location of the referent.
3) \textbf{RefCOCOg}~\cite{mao2016generation}: It consists of 104,560 referring expressions for 54,822 objects in 26,711 images. These expressions are collected in a non-interactive way, and the average length of expression is 8.4 words. Compared to RefCOCO and RefCOCO+, expressions in RefCOCOg are more descriptive and involve more objects. We follow the same split as~\cite{nagaraja2016modeling}.

\textbf{Evaluation Metrics.} We evaluated VL-NMS on two different REG tasks: referring expression comprehension (REC) and referring expression segmentation (RES). For the REC task, we used top-1 accuracy as the evaluation metric. When the IoU between the predicted and ground truth is larger than 0.5, the prediction is considered to be correct. For the RES task, we used the overall IoU and Pr@X (the percentage of samples with IoU higher than X)\footnote{Due to the limited space, all RES results with the Pr@X metric are provided in the supplementary materials.} as metrics. \cc{Besides, we calculated the average performance on the three benchmarks to show the global performance of the model.}

\addtolength{\tabcolsep}{-2pt}
\begin{table*}[t]
	\small
	\caption{Recall (\%) of the referent and contextual objects. The baseline detector is a ResNet-101 based Mask R-CNN with plain GreedyNMS. `B' denotes VL-NMS with binary XE loss, `R' denotes VL-NMS with ranking loss. `Real' denotes the real case used in the state-of-the-art two-stage methods.}
	\vspace{-7pt}
    \begin{center}
		\scalebox{0.95}{
			\begin{tabular}{ c c c c c c c c c c}
				\toprule
				\multicolumn{10}{c}{Referent} \\
				\midrule
				&\multirow{2.5}{*}{VL-NMS$_\text{FC}$}  & \multicolumn{3}{c}{RefCOCO} & \multicolumn{3}{c}{RefCOCO+} & \multicolumn{2}{c}{RefCOCOg} \\
				\cmidrule(lr){3-5} \cmidrule(lr){6-8} \cmidrule(lr){9-10}
				& & val & testA & testB &  val & testA & testB & val & test \\
				\midrule
				\parbox[t]{2mm}{\multirow{3}{*}{\rotatebox[origin=c]{90}{N=100}}} &  & 97.60 & 97.81 & 96.58 & 97.79 & 97.78 & 96.99 & 97.18 & 96.91 \\
				& B & 97.75 & 98.59 & 97.08 & 97.96 & 98.39 & 97.50 & 97.61 & 97.44  \\
				& R & 97.62 & 98.02 & 96.78 & 97.71 & 98.06 & 97.14 & 97.18 & 97.08 \\
				\midrule
				\parbox[t]{2mm}{\multirow{3}{*}{\rotatebox[origin=c]{90}{Real}}} &  & 88.84 & 93.99 & 80.77 & 90.71 & 94.34 & 84.11 & 87.83 & 87.88 \\
				& B & \textbf{92.51} & \textbf{95.56} & \textbf{88.28} & \textbf{93.42} & \textbf{95.86} & \textbf{88.95} & \textbf{90.28} & \textbf{90.34} \\
				& R & 90.50 & 94.75 & 83.87 & 91.62 & 95.14 & 86.42 & 89.01 & 88.96 \\
				\midrule
				\multicolumn{10}{c}{Contextual Objects} \\
				\midrule
				&\multirow{2.5}{*}{VL-NMS$_\text{FC}$} & \multicolumn{3}{c}{RefCOCO} & \multicolumn{3}{c}{RefCOCO+} & \multicolumn{2}{c}{RefCOCOg} \\
				\cmidrule(lr){3-5} \cmidrule(lr){6-8} \cmidrule(lr){9-10}
				&  & val & testA & testB &  val & testA & testB & val & test \\
				\midrule
				\parbox[t]{2mm}{\multirow{3}{*}{\rotatebox[origin=c]{90}{N=100}}} &  & 90.14 & 89.85 & 90.53 & 89.53 & 88.47 & 90.69 & 90.56 & 90.30 \\
				& B & 90.38 & 90.31 & 90.64 & 89.67 & 88.88 & 91.04 & 90.36 & 90.37 \\
				& R  & 90.22 & 89.83 & 90.63 & 89.70 & 88.62 & 90.71 & 90.67 & 90.30 \\
				\midrule
				\parbox[t]{2mm}{\multirow{3}{*}{\rotatebox[origin=c]{90}{Real}}} &  &74.97 & 78.60 & 70.19 & 76.34 & 77.45 & 73.52 & 75.69 & 75.87 \\
				& B & \textbf{78.75} & \textbf{80.14} & \textbf{76.47} & \textbf{78.44} & \textbf{78.82} & \textbf{77.49} & 76.12 & 76.57 \\
				& R & 76.79 & 79.12 & 72.99 & 77.66 & 78.44 & 75.59 & \textbf{76.68} & \textbf{76.73}  \\
				\bottomrule
			\end{tabular}
		} 
	\end{center}
	\label{tab:recall}
    \vspace{-3pt}
\end{table*}
\addtolength{\tabcolsep}{2pt}

\addtolength{\tabcolsep}{-0pt}    
\begin{table*}[t]
	\small
	\caption{Performance of VL-NMS on REC and RES with different grounding backbones. The metric is top-1 accuracy (\%) for REC and overall IoU (\%) for RES. All baselines use the ResNet-101 based Mask R-CNN as first-stage networks. The best and second best methods under each setting are marked in bold and italic fonts, respectively. $^\dagger$~denotes the results are from our implementations.}
    \vspace{-7pt}	
    \begin{center}
		\scalebox{0.95}{
			\begin{tabular}{ l c c c c c c c c }
				\toprule
				\multirow{2.5}{*}{Models} & \multicolumn{3}{c}{RefCOCO} & \multicolumn{3}{c}{RefCOCO+} & \multicolumn{2}{c}{RefCOCOg} \\
				\cmidrule(lr){2-4} \cmidrule(lr){5-7} \cmidrule(lr){8-9} 
				& val & testA & testB &  val & testA & testB & val & test \\
				\midrule
				\multicolumn{9}{l}{Referring Expression Comprehension} \\
			    \midrule
				MAttNet~\cite{yu2018mattnet}  & 76.65 & 81.14 & 69.99 & 65.33 & 71.62 & 56.02 & 66.58 & 67.27 \\
				MAttNet$^\dagger$ & 76.92 & 81.19 & 69.58 & 65.90 & \textit{71.53} & 57.23 & 67.52 & 67.55  \\
				~~+VL-NMS$_\text{FC}$ B & \textit{78.82} & \textbf{82.71} & \textit{73.94} & \textbf{66.95} & 71.29 & \textit{58.40} & \textit{68.89} & \textit{68.67} \\
				~~+VL-NMS$_\text{FC}$ R & 77.98 & 82.02 & 71.64 & \textit{66.64} & 71.36 & 58.01 & \textbf{69.16} & 67.63 \\
				~~+VL-NMS$_\text{Trans}$ & \textbf{79.16} & \textit{82.31} & \textbf{74.52} & 66.58 & \textbf{71.85} & \textbf{58.48} & 68.87 & \textbf{68.75} \\
				\midrule
				NMTree~\cite{liu2019learning} & 76.41 & 81.21 & 70.09 & 66.46 & 72.02 & 57.52 & 65.87 & 66.44 \\
				NMTree$^\dagger$ & 76.13 & 80.17 & 70.19 & 66.65 & 71.48 & 57.74 & 65.65 & 65.94 \\
				~~+VL-NMS$_\text{FC}$ B & 77.39 & \textit{81.40} & \textit{72.46} & \textbf{67.15} & \textit{71.76} & 58.70 & \textit{67.30} & \textbf{66.93} \\
				~~+VL-NMS$_\text{FC}$ R &  77.81 & \textbf{81.69} & 71.78 & \textit{67.03} & \textbf{71.78} & \textit{58.79} & 66.81 & 66.31 \\
				~~+VL-NMS$_\text{Trans}$ & \textbf{78.00} & 81.09 & \textbf{72.93} & 66.67 & 71.55 & \textbf{58.85} & \textbf{67.83} & \textit{66.87} \\
				\midrule
				CM-A-E~\cite{liu2019improving}  & 78.35 & 83.14 & 71.32 & 68.09 & 73.65 &	58.03 & 67.99 & 68.67 \\
				CM-A-E$^\dagger$ & 78.35 & 83.12 & 71.32 & 68.19 & 73.04 & 58.27 & 69.10 & 69.20 \\
				~~+VL-NMS$_\text{FC}$ B & \textit{80.70} & \textit{84.00} & \textit{76.04} & 68.25 & \textbf{73.68} & \textit{59.42} & \textit{70.55} & \textbf{70.62} \\
				~~+VL-NMS$_\text{FC}$ R & 79.55 & \textit{83.58} & 73.62 & \textit{68.51} & 73.14 & 58.38 & 69.77 & 70.01 \\
				~~+VL-NMS$_\text{Trans}$ & \textbf{80.97} & \textbf{84.02} & \textbf{76.64} & \textbf{68.76} & \textit{73.30} & \textbf{60.44} & \textbf{70.81} & \textit{70.30} \\
				\midrule
				\multicolumn{9}{l}{Referring Expression Segmentation} \\
				 \midrule
				 MAttNet~\cite{yu2018mattnet} & 56.51 & 62.37 & 51.70 & 46.67 & 52.39 & 40.08 & 47.64 & 48.61 \\
				 MAttNet$^\dagger$ & 57.14  &  62.34  & 51.48  & 47.30  & \textbf{52.37} & 41.14 & 48.28  & 49.01 \\
				 ~~+VL-NMS$_\text{FC}$ B & \textbf{59.75} & \textbf{63.48} & \textit{55.66} & \textbf{48.39} & 51.57 & \textbf{42.56} & 49.54 & \textit{50.38} \\
				 ~~+VL-NMS$_\text{FC}$ R & 58.32 & 62.96 & 53.68 & 47.87 & 51.85 & 41.41 & \textbf{50.13} & 49.07 \\
				 ~~+VL-NMS$_\text{Trans}$ & \textit{59.73} & \textit{63.43} & \textbf{56.44} & \textit{48.32} & \textit{52.00} & \textit{42.27} & \textit{49.81} & \textbf{50.67} \\
				 \midrule
				 NMTree~\cite{liu2019learning} & 56.59 & 63.02 & 52.06 & 47.40 & 53.01 & 41.56 & 46.59 & 47.88 \\
				 NMTree$^\dagger$ & 56.78 & 61.83 & 52.94 & 47.75 & \textit{52.36} & 41.86 & 46.19 & 47.41 \\
				 ~~+VL-NMS$_\text{FC}$ B & 58.35 & \textit{62.59} & \textbf{55.40} & \textbf{48.68} & 52.30 & \textit{42.64} & \textit{48.14} & \textit{48.59} \\
				 ~~+VL-NMS$_\text{FC}$ R & \textit{58.42} & \textbf{62.69} & 53.60 & 48.27 & \textbf{52.65} & 42.18 & 47.72 & 48.09 \\
				 ~~+VL-NMS$_\text{Trans}$ & \textbf{59.26} & 62.25 & \textit{55.00} & \textit{48.31} & 52.09 & \textbf{42.71} & \textbf{49.09} & \textbf{48.67} \\
				 \midrule
				 CM-A-E~\cite{liu2019improving} & --- & --- & --- & --- & --- & --- & --- & --- \\
				 CM-A-E$^\dagger$ & 58.23  & 64.60 & 53.14  & 49.65  & \textbf{53.90}  & 41.77  & 49.10  & 50.72 \\
				 ~~+VL-NMS$_\text{FC}$ B & \textit{61.46} & \textbf{65.55} & \textit{57.41} & 49.76 & \textit{53.84} & \textit{42.66} & \textit{51.21} & \textbf{51.90} \\
				 ~~+VL-NMS$_\text{FC}$ R & 59.72 & 64.87 & 55.63 & \textit{49.86} & 52.62 & 41.87 & 50.13 & 51.44 \\
				 ~~+VL-NMS$_\text{Trans}$  & \textbf{61.72} & \textit{65.23} & \textbf{58.37} & \textbf{50.33} & \textbf{53.90} & \textbf{43.98} & \textbf{51.25} & \textit{51.74} \\
				\bottomrule
			\end{tabular}
		} 
	\end{center}
	\label{tab:model-agnostic}
 \vspace{-3pt}
\end{table*}
\addtolength{\tabcolsep}{0pt}

\textbf{Implementation Details.} We built a vocabulary for each dataset by filtering out the words less than 2 times, and exploited the 300-d GloVe embeddings as the initialization of word embeddings. We used an ``unk" symbol to replace all words out of the vocabulary. The largest length of sentences was set to 10 for RefCOCO and RefCOCO+, 20 for RefCOCOg. The text encoder $e_q$ is a bidirectional GRU with a hidden size of 256. For the visual encoder $e_v$, we used the same head network of the Mask R-CNN with ResNet-101 backbone\footnote{https://github.com/lichengunc/mask-faster-rcnn} as prior works~\cite{yu2018mattnet}, and utilized the pre-trained weights as initialization. The weights of the original detector (\ie, the gray part in Figure~\ref{fig:architecture}) were fixed during training. For VL-NMS$_{\text{Trans}}$, the largest length of sentences was doubled and we applied BERT as text encoder following~\cite{lu2019vilbert}. We used the 2-layer Transformer and co-attention network to provide interaction between two modalities. The whole model was trained with the Adam optimizer. The learning rate was initialized to 4e-4 for the head network, 2e-6 for the relatedness module of VL-NMS$_{\text{Trans}}$ and 5e-3 for the rest of network. We used a batch size of 64 for VL-NMS$_{\text{Trans}}$ and 8 for VL-NMS$_{\text{FC}}$. The thresholds $\delta$ and $\gamma$ were set to 0.05 and 0.4, respectively. For the ranking loss, the top-h was set to 100. We built pseudo ground-truth based on text similarity, if not mentioned otherwise. 

\subsubsection{Recall Analyses of Critical Objects} To evaluate the effectiveness of the VL-NMS in improving the recall of both referent and contextual objects, we compared VL-NMS with plain GreedyNMS used in the baseline detector (\ie, ResNet-101 based Mask R-CNN). Since we only have annotated ground-truth bboxes for the referent, we calculated the recall of pseudo ground-truths to approximate the recall of contextual objects. The results are reported in Table~\ref{tab:recall}, and more detailed results are provided in the supplementary materials.
From Table~\ref{tab:recall}, we have the following observations. When using top-100 bboxes as proposals, all three methods can achieve near-perfect recall ($\approx$ 97\%) for the referent and acceptable recall ($\approx$ 90\%) for the contextual objects, respectively. However, as the number of proposals decreases to a very small number (\eg, $<$ 10 in the real case), the recall of the baseline all drops significantly (\eg, 15.81\% for the referent and 20.34\% for the contextual objects on RefCOCO testB). In contrast, VL-NMS can help narrow the gap over all dataset splits. Especially, the improvement is more obvious on the testB set (\eg, 7.51\% and 4.85\% absolute gains for the recall of referent on RefCOCO and RefCOCO+), where the categories of referents are more diverse and the recalls are relatively lower.

\subsubsection{Architecture Agnostic Generalization} Since the VL-NMS model is agnostic to the second stage network, it can be easily integrated into any referent grounding architecture. To evaluate the effectiveness and generalizability of VL-NMS in boosting the performance of different backbones, we incorporated VL-NMS into three state-of-the-art two-stage grounding methods: \textbf{MAttNet}~\cite{yu2018mattnet} , \textbf{NMTree}~\cite{liu2019learning}, and \textbf{CM-A-E}~\cite{liu2019improving}. All results are reported in Table~\ref{tab:model-agnostic}.
From Table~\ref{tab:model-agnostic}, we can observe that all variants of VL-NMS can consistently improve the grounding performance of three backbones on both REC and RES. The improvement is more significant on the testB set (\eg, 5.32\% and 5.23\% absolute performance gains for CM-A-E in REC and RES), which meets our expectation, \ie, the improvements in the recall of critical objects at the first stage have a strong positive correlation with the improvements of grounding performance at the second stage. Comparing the two variants of VL-NMS, in most of the cases, VL-NMS$_{\text{Trans}}$ achieves better grounding performance.

\begin{table}[t]
	\small
	\caption{Comparison with state-of-the-art models on REC and RES. The metric is top-1 accuracy (\%) for REC and overall IoU (\%) for RES. $^\dagger$ denotes the results are from our implementation.}
    \vspace{-7pt}	
    \setlength{\tabcolsep}{1.7mm}	
    \begin{center}
		\scalebox{0.90}{
			\begin{tabular}{l  l  c  c c  c c  c c}
				\toprule
				& \multirow{2}{*}{Models} & \multirow{2}{*}{Backbone} & \multicolumn{2}{c}{RefCOCO} & \multicolumn{2}{c}{RefCOCO+} & RefCOCOg & \multirow{2}{*}{\cc{Mean}}\\
				& & & testA & testB & testA & testB & test & \\
				\midrule
				\multicolumn{9}{l}{Referring Expression Comprehension} \\
				\midrule
				\parbox[t]{0mm}{\multirow{4}{*}{\rotatebox[origin=c]{90}{one-s.}}} & SSG~\cite{chen2018real} & darknet53 & 76.51 & 67.50 & 62.14 & 49.27 & --- & \cc{---}\\  
				& FAOA~\cite{yang2019fast} & darknet53 & 74.88 & 66.32 & 61.89 & 49.46 & 58.90 & \cc{62.29} \\
				& RCCF~\cite{liao2020real} & dla34 & 81.06 & 71.85 & 70.35 & 56.32 & 65.73 & \cc{69.06} \\
				& RSC-Large~\cite{yang2020improving} & darknet53 & 80.45 & 72.30 & 68.36 & 67.30 & 67.20 & \cc{71.12} \\
				\midrule
                \parbox[t]{0mm}{\multirow{5}{*}{\rotatebox[origin=c]{90}{\cc{transformer}}}} 
                & \cc{TransVG~\cite{deng2021transvg}} & \cc{res101} & \cc{83.38} & \cc{76.94} & \cc{72.46} & \cc{59.24} & \cc{67.98} & \cc{72.00} \\
				& \cc{QRNet~\cite{ye2022shifting}} & \cc{Swin-S} & \cc{85.85} & \cc{82.34} & \cc{76.17} & \cc{63.81} & \cc{72.52} & \cc{76.14} \\
				& \cc{VLTVG~\cite{yang2022improving}} & \cc{res101} & \cc{87.24} & \cc{80.49} & \cc{78.93} & \cc{65.17} & \cc{74.18} & \cc{77.20} \\
                & \cc{SeqTR~\cite{zhu2022seqtr}} & \cc{darknet53} & \cc{86.51} & \cc{81.24} & \cc{76.26} & \cc{64.88} & \cc{74.21} & \cc{76.62} \\  
                & \cc{SiRi~\cite{qu2022siri}} & \cc{res101} & \cc{88.56} & \cc{81.27} & \cc{82.01} & \cc{66.33} & \cc{76.46} & \cc{78.93} \\
				\midrule
				\parbox[t]{1mm}{\multirow{11}{*}{\rotatebox[origin=c]{90}{two-s.}}} & VC~\cite{zhang2018grounding} & vgg16 & 73.33 & 67.44 & 58.40 & 53.18 & --- & \cc{---}\\
				& ParalAttn~\cite{zhuang2018parallel} & vgg16 & 75.31 & 65.52 & 61.34 & 50.86 & --- & \cc{---}\\
				& LGRANs~\cite{wang2019neighbourhood} & vgg16 & 76.60 & 66.40 & 64.00 & 53.40 & --- & \cc{---}\\
				& DGA~\cite{yang2019dynamic} & vgg16 & 78.42 & 65.53 & 69.07 & 51.99 & 63.28 & \cc{65.66} \\ 
				& NMTree~\cite{liu2019learning} & vgg16 & 74.81 & 67.34 & 61.09 & 53.45 & 61.46 & \cc{63.63} \\
				& MAttNet~\cite{yu2018mattnet} & res101 & 81.14 & 69.99 & 71.62 & 56.02 & 67.27 & \cc{69.21} \\
				& RvG-Tree~\cite{hong2019learning} & res101 & 78.61 & 69.85 & 67.45 & 56.66 & 66.51 & \cc{67.82} \\
				& NMTree~\cite{liu2019learning} & res101 & 81.21 & 70.09 & 72.02 & 57.52 & 66.44 & \cc{69.45} \\
				& CM-A-E~\cite{liu2019improving} & res101 &	83.14 & 71.32 &	73.65 &	58.03 & 68.67 & \cc{70.96} \\ 
				& CM-A-E+VL-NMS$_\text{FC}$ & res101 & 84.00 & 76.04 & \textbf{73.68} & 59.42 & \textbf{70.62} & \cc{72.75} \\
				& \textbf{CM-A-E+VL-NMS$_{\text{Trans}}$} & res101 & \textbf{84.02} & \textbf{76.64} & 73.30 & \textbf{60.44} & 70.30 & \textbf{\cc{72.94}} \\
				\midrule
				\multicolumn{9}{l}{Referring Expression Segmentation} \\
				\midrule
				\parbox[t]{1mm}{\multirow{6}{*}{\rotatebox[origin=c]{90}{one-s.}}} & STEP~\cite{chen2019see} & res101 & 63.46 & 57.97 & 52.33 & 40.41 & --- & \cc{---}\\
				& BRINet~\cite{hu2020bi} & res101 & 62.99 & 59.21 & 52.32 & 42.41 & --- & \cc{---}\\
				& CMPC~\cite{huang2020referring} & res101 & 64.53 & 59.64 & 53.44 & 43.23 & --- & \cc{---}\\
				& MCN~\cite{luo2020multi} & darknet53 & 64.20 & 59.71 & 54.99 & 44.69 & 49.40 & \cc{54.60} \\
                & \cc{LTS~\cite{jing2021locate}} & \cc{darknet53} & \cc{67.76} & \cc{63.08} & \cc{58.32} & \cc{48.02} & \cc{54.25} & \cc{58.27} \\
                & \cc{VLT~\cite{ding2021vision}} & \cc{darknet56} & \cc{68.29} & \cc{62.73} & \cc{59.20} & \cc{49.36} & \cc{56.65} & \cc{59.24} \\
				\midrule
				\parbox[t]{1mm}{\multirow{5}{*}{\rotatebox[origin=c]{90}{two-s.}}} & MAttNet~\cite{yu2018mattnet} & res101 & 62.37 & 51.70 & 52.39 & 40.08 & 48.61 & \cc{51.03} \\
				& NMTree~\cite{liu2019learning} & res101 & 63.02 & 52.06 & 53.01 & 41.56 & 47.88 & \cc{51.51} \\
				& CM-A-E$^\dagger$~\cite{liu2019improving} & res101 & 64.60 & 53.14 & \textbf{53.90}  & 41.77 & 50.72 & \cc{52.83} \\
				& CM-A-E+VL-NMS$_\text{FC}$ & res101 & \textbf{65.55} & 57.41 & 53.84 & 42.66 & \textbf{51.90} & \cc{54.27} \\ 
				& \textbf{CM-A-E+VL-NMS$_{\text{Trans}}$} & res101 & 65.23 & \textbf{58.37} & \textbf{53.90} & \textbf{43.98} & 51.74 & \textbf{\cc{54.65}} \\
				\bottomrule
			\end{tabular}
		} 
	\end{center}
	\label{tab:sota}
    \vspace{-3pt}
\end{table}

\subsubsection{Comparison with State-of-the-Arts}

We incorporate VL-NMS$_\text{FC}$ and VL-NMS$_{\text{Trans}}$ into CM-A-E model, dubbed \textbf{CM-A-E+VL-NMS$_\text{FC}$} and \textbf{CM-A-E+VL-NMS$_{\text{Trans}}$}, and compare them against state-of-the-art REC and RES methods. For fair comparison, we group state-of-the-art REC methods into: 1) two-stage methods: \textbf{VC}~\cite{zhang2018grounding}, \textbf{ParalAttn}~\cite{zhuang2018parallel}, \textbf{LGRANs}~\cite{wang2019neighbourhood}, \textbf{DGA}~\cite{yang2019dynamic}, \textbf{NMTree}~\cite{liu2019learning}, \textbf{MAttNet}~\cite{yu2018mattnet}, \textbf{RvG-Tree}~\cite{hong2019learning}, and \textbf{CM-A-E}~\cite{liu2019improving}; 2) one-stage methods: \textbf{SSG}~\cite{chen2018real}, \textbf{FAOA}~\cite{yang2019fast}, \textbf{RCCF}~\cite{liao2020real}, and \textbf{RSC-Large}~\cite{yang2020improving}; \cc{3) Transformer-based methods: \textbf{TransVG}\cite{deng2021transvg}, \textbf{SeqTR}\cite{zhu2022seqtr}, \textbf{QRNet}\cite{ye2022shifting}, \textbf{VLTVG}\cite{yang2022improving} and \textbf{SiRi}\cite{qu2022siri}}. Analogously, we group state-of-the-art RES methods into: 1) two-stage methods: \textbf{MAttNet}~\cite{yu2018mattnet}, \textbf{NMTree}~\cite{liu2019learning}, and \textbf{CM-A-E}~\cite{liu2019improving}; 2) one-stage methods: \textbf{STEP}~\cite{chen2019see}, \textbf{BRINet}~\cite{hu2020bi}, \textbf{CMPC}~\cite{huang2020referring}, \textbf{MCN}~\cite{luo2020multi}, \cc{\textbf{LTS}~\cite{jing2021locate}, and \textbf{VLT}~\cite{ding2021vision}.}
The results are reported in Table~\ref{tab:sota}. For the REC task, CM-A-E+VL-NMS achieves a new record-breaking performance that is superior to all existing two-stage REC methods on three benchmarks. Specifically, VL-NMS improves the strong baseline CM-A-E with an average of 2.94\%, 0.91\%, 2.23\% absolute performance gains on RefCOCO, RefCOCO+, and RefCOCOg. For the RES task, CM-A-E+VL-NMS achieves a new state-of-the-art performance of two-stage methods over most of the dataset splits. Similarly, VL-NMS improves CM-A-E with an average of 3.12\%. 0.96\%, 1.59\% absolute performance gains on three benchmarks. \cc{Meanwhile, we calculated the mean performance on three benchmarks to intuitively explain the global effect of our method. The results have reported in Table~\ref{tab:sota}, denoted by "\textit{Mean}". It can be seen that the improvement of the new proposed VL-NMS$_{\text{Trans}}$ is approximately the same as that between NMTree and MAttNet, two previous state-of-the-art methods. According to Table~\ref{tab:sota}, Transformer-based models tend to have better performance due to larger parameter sizes and end-to-end training. But in our setting, we focus on comparing two-stage methods because they typically have better interpretability and generalization.} Note that since one-stage and two-stage RES models are pretrained on different datasets, the comparison between them is not strictly fair.

\begin{table}[t]
	\small
	\caption{Ablation studies of text similarity based pseudo ground-truth on REC task. The grounding backbone is CM-A-E.}
    \vspace{-7pt}
	\begin{center}
		\scalebox{0.95}{
			\begin{tabular}{ c c c c c c}
				\toprule
				\multirow{2}{*}{VL-NMS} & \multicolumn{2}{c}{RefCOCO} & \multicolumn{2}{c}{RefCOCO+} & \multicolumn{1}{c}{RefCOCOg}  \\
				& testA & testB & testA & testB & test  \\
				\midrule
				w/o Pseudo GT & 83.63 & \textbf{76.09} & 71.83 & 58.64 & 68.76 \\
				w/ Pseudo GT & \textbf{84.00} & 76.04 & \textbf{73.68} & \textbf{59.42} & \textbf{70.62} \\
				\bottomrule
			\end{tabular}
		} 
	\end{center}
	\label{tab:multi-gt}
    \vspace{-7pt}
\end{table}

\subsubsection{Ablation Studies of Pseudo Ground-truth}

To validate the effectiveness of pseudo ground-truth, we further compare VL-NMS with a strong baseline: VL-NMS without pseudo ground-truth. The results are shown in Table~\ref{tab:multi-gt}. Both methods are trained with the same set of hyper-parameters and tested with CM-A-E as the grounding backbone. We can observe that VL-NMS performs better on all splits except RefCOCO testB where the performance difference between the two methods is trivial (0.05\%). Especially on the RefCOCOg dataset where the expressions are more complex, training with pseudo ground-truth lifts the performance by 1.8\%. These results validate the effectiveness of using pseudo ground-truth for contextual objects in the training phase. Meanwhile, from another perspective, VL-NMS trained without pseudo ground-truth can be regarded as a "generalized" one-stage REC model. This suggests that current one-stage REC methods, trained to yield the referent instead of all critical objects, are typically not as qualified as VL-NMS for generating proposals for two-stage REG methods.

\subsubsection{Grounding without COCO Annotations}
As all three REG benchmarks are built upon the COCO dataset, most state-of-the-art grounding methods, either explicitly or implicitly, make use of extra COCO annotations. Two-stage methods normally use a COCO pretrained detector to extract proposals at the first stage while one-stage methods utilize the pretrained detector weights as initialization of backbone networks. In VL-NMS, we also use COCO annotations to build pseudo ground-truth for contextual objects by the text similarity based method. To relieve the two-stage REG framework from depending on extra data, we attempt to remove COCO detection annotations (\ie, ground-truth bboxes and categories) from the grounding pipeline.
During training, we replaced the COCO annotations with the detected bboxes and categories by the widely used Visual Genome pretrained detector~\cite{anderson2018bottom}. Then, we built pseudo ground-truth for contextual objects in a similar way to text similarity based method. We grounded all noun chunks in the expression, so it can be interpreted as a WSPG method.
Then we regarded all pseudo ground-truth bboxes and the annotated referent bbox as foreground bboxes to train VL-NMS. Since only the ground-truth has credible annotations, we merely performed regression on them. Meanwhile, we empirically reduce the weight and label of pseudo ground-truth to 0.5. In the testing phase, we also replace the COCO detector with the Visual Genome detector. The results are shown in Table~\ref{tab:rec-bottom-up}. We can observe that replacing the COCO pretrained detector results in a much inferior grounding performance compared to Table~\ref{tab:model-agnostic} (\eg, performance of CM-A-E decreases by 35.94\%). Applying VL-NMS with WSPG-based pseudo ground-truth can consistently and greatly boost the performance of all baselines over three REG benchmarks (\eg, performance of CM-A-E increases by 32.27\%, almost as high as the original CM-A-E+VL-NMS in Table~\ref{tab:model-agnostic}). We also compared the two ways of generating pseudo ground-truth. As expected, text similarity based method (dubbed VL-NMS (COCO)) generally yields superior results over WSPG based methods (dubbed VL-NMS (WSPG)) by exploiting COCO annotations. \cc{The gap is most obvious on RefCOCOg, indicating that it is most sensitive to the quality of the contextual objects. Since its expressions have the longest average lengths and contain the most contextual objects.} But the performance gap between these two methods is negligible.

\subsubsection{Inference Time of VL-NMS}
As VL-NMS needs to forward the detected bboxes of a Mask RCNN back through the RoI pooling layer to calculate the relatedness score, it is reasonable to concern about the added time complexity of VL-NMS. We quantitatively measured the inference time of each component of the two-stage grounding pipeline with VL-NMS. As shown in Table~\ref{tab:time}, the inference time of VL-NMS is marginal ($\approx$ 16\% extra time for both MAttNet and NMTree).

\begin{table}[t]
	\small
	\caption{Effectiveness of weakly supervised phrase grounding based pseudo ground-truth on REC task. To avoid using COCO annotations, all baselines use Faster R-CNN pretrained on Visual Genome dataset~\cite{anderson2018bottom} as first-stage networks. The best and second best methods under each setting are marked in bold and italic fonts, respectively.}
    \vspace{-7pt}
	\setlength{\tabcolsep}{1.7mm}
	\begin{center}
		\scalebox{0.90}{
			\begin{tabular}{ l  c c  c c c }
				\toprule
				\multirow{2}{*}{Models} & \multicolumn{2}{c}{RefCOCO} & \multicolumn{2}{c}{RefCOCO+} & RefCOCOg \\
				& testA & testB & testA & testB & test \\
				\midrule
				MAttNet~\cite{yu2018mattnet} & 45.87 & 51.64 & 45.60 & 44.45 & 45.33 \\
				~~+VL-NMS$_\text{FC}$ (WSPG) & 76.97 & 68.15 & 67.95 & \textit{55.86} & \textit{64.32} \\
				~~+VL-NMS$_{\text{Trans}}$ (WSPG) & \textbf{78.47} & \textbf{72.25} & \textit{68.16} & 54.90 & 63.63 \\
				~~+VL-NMS$_\text{FC}$ (COCO) & \textit{77.92} & \textit{70.70} & \textbf{68.93} & \textbf{56.33} & \textbf{64.85} \\
				\midrule
				NMTree~\cite{liu2019learning} & 46.72 & 51.03 & 46.02 & 44.43 & 44.59 \\
				~~+VL-NMS$_\text{FC}$ (WSPG) & 75.84 & 66.63 & 68.37 & \textbf{55.16} & \textit{63.80} \\
				~~+VL-NMS$_{\text{Trans}}$ (WSPG) &  \textbf{78.58} & \textbf{71.03} & \textbf{69.49} & \textit{55.14} & 63.38\\
				~~{+VL-NMS$_\text{FC}$ (COCO)} & \textit{77.43} & \textit{69.28} & \textit{69.46} & 55.12 & \textbf{64.26} \\
				\midrule
				CM-A-E~\cite{liu2019improving} & 48.06 & 52.74 & 48.01 & 45.76 & 46.29 \\
				~~+VL-NMS$_\text{FC}$ (WSPG) & 79.28 & 69.89 & 69.28 & \textbf{57.76} & \textit{65.87} \\
				~~+VL-NMS$_{\text{Trans}}$ (WSPG) & \textbf{80.33} & \textbf{73.03} & \textbf{70.31} & \textit{56.31} & 65.36 \\
				~~{+VL-NMS$_\text{FC}$ (COCO)} & \textit{80.06} & \textit{72.68} & \textit{70.24} & 56.29 & \textbf{66.03} \\
				\bottomrule
			\end{tabular}
		} 
	\end{center}
	\label{tab:rec-bottom-up}
    \vspace{-7pt}
\end{table}

\begin{table}[t]
	\small
	\caption{Inference time (seconds) of two-stage grounding pipeline with VL-NMS. Measured on a single NVIDIA 1080Ti.}
    \vspace{-7pt}
	\begin{center}
		\scalebox{0.95}{
			\begin{tabular}{c c c c}
				\toprule
				\multicolumn{2}{c}{First Stage} & \multicolumn{2}{c}{Second Stage} \\
				\cmidrule(lr){1-2} \cmidrule(lr){3-4}
				Res101-MRCNN & \textbf{VL-NMS$_\text{FC}$} & MAttNet & NMTree \\
				0.1893 & 0.0974 & 0.4002 & 0.4058 \\
				\bottomrule
			\end{tabular}
		}
	\end{center}
	\label{tab:time}
    \vspace{-7pt}
\end{table}

\begin{table}[t]
	\small
	\caption{Performance of VL-NMS$_\text{FC}$ and VL-NMS$_{\text{Trans}}$ on phrase grounding. the metric is top-1 accuracy (\%). the best methods are marked in bold.}
    \vspace{-7pt}
	\begin{center}
		\setlength{\tabcolsep}{4mm}{\scalebox{0.95}{
			\begin{tabular}{ l l c }
				\toprule
				\multicolumn{2}{c}{Models} & Accuracy(\%) \\
				\midrule
				\parbox[t]{1mm}{\multirow{3}{*}{\rotatebox[origin=c]{90}{one-s.}}}
				& ZSGNet~\cite{sadhu2019zero} & 63.39 \\
				& FAOA~\cite{yang2019fast} & 68.71 \\
				& RSC-Large~\cite{yang2020improving} & 69.28 \\
				\midrule
				\parbox[t]{1mm}{\multirow{5}{*}{\rotatebox[origin=c]{90}{two-s.}}}
				& Similarity Net \cite{wang2018learning} & 60.89 \\
				& CITE \cite{plummer2018conditional} & 61.33 \\
				& DDPN \cite{yu2018rethinking} & 73.30 \\
				& DDPN+VL-NMS$_\text{FC}$ & 73.59 \\
				& \textbf{DDPN+VL-NMS$_{\text{Trans}}$} & \textbf{73.99}\\
				\bottomrule
			\end{tabular}
		}}
	\end{center}
	\label{tab:phrase grounding}
    \vspace{-7pt}
\end{table}

\subsection{Results on Phrase Grounding}

\textbf{Datasets.} We evaluated VL-NMS on a commonly used phrase grounding benchmark: \textbf{Flickr-30K Entities}~\cite{plummer2015flickr30k}. It consists of 31,783 images and 427K referred entities. We split 1,000 images for validation, 1,000 images for test, and the rest for training following previous work~\cite{yu2018rethinking}.

\textbf{Evaluation Metrics.} We adopted accuracy as the evaluation metric. The prediction is considered to be correct if the IoU between the prediction and ground truth is larger than 0.5.

\textbf{Baselines.} We applied \textbf{DDPN} \cite{yu2018rethinking} as the backbone and integrated VL-NMS into it, as DDPN is the state-of-the-art method for two-stage phrase grounding.

su\textbf{Results.} Table \ref{tab:phrase grounding} reports the results of VL-NMS on the Flickr30K Entities test set. Compared with the state-of-the-art two-stage method DDPN, the performance improved by 0.69\%, which proves that VL-NMS can be well transferred to the phrase grounding task.

\begin{table*}[t]
	\small
	\caption{Performance of VL-NMS on ITM task with different matching backbones on Flickr30K. The metrics are Recall@K (\%) and Recall@Sum(\%). The best and second best methods under each setting are marked in bold and italic fonts, respectively.}
    \vspace{-7pt}
	\begin{center}
		\scalebox{0.95}{
			\begin{tabular}{l l  c c c  c c c  c}
				\toprule
				& \multirow{2}{*}{Models} & \multicolumn{3}{c}{Sentence Retrieval} & \multicolumn{3}{c}{Image Retrieval} & \multirow{2}{*}{R@sum} \\
				& & R@1 & R@5 & R@10 & R@1 & R@5 & R@10 \\
				\midrule
				\parbox[t]{2mm}{\multirow{8}{*}{\rotatebox[origin=c]{90}{N=10}}} & SCAN~\cite{lee2018stacked} & 57.8 & 83.1 & 90.1 & 40.8 & 68.6 & 78.1 & 418.5 \\
				& ~~+VL-NMS$_\text{FC}$ (GT) & \textit{65.4} & \textit{86.6} & \textit{91.9} & \textit{42.9} & \textit{71.0} & \textit{79.7} & \textit{437.5} \\
				& ~~+VL-NMS$_\text{FC}$ (Transfer) & 61.1 & 86.6 & 91.5 & 41.7 & 70.5 & \textbf{79.8} & 431.2 \\
				& ~~+VL-NMS$_{\text{Trans}}$ (GT) & \textbf{68.1} & \textbf{88.9} & \textbf{93.4} & \textbf{46.7} & \textbf{71.3} & 79.1 & \textbf{447.5} \\
				\cmidrule(lr){2-9}
				& IMRAM~\cite{chen2020imram} & 49.8 & 78.3 & 87.5 & 34.2 & 62.9 & 72.9 & 385.6 \\
				& ~~+VL-NMS$_\text{FC}$ (GT) & \textit{67.8} & \textit{87.6} & \textit{93.0} & \textit{44.5} & \textbf{71.7} & \textit{80.1} & \textit{444.7} \\
				& ~~+VL-NMS$_\text{FC}$ (Transfer) & 59.9 & 85.2 & 91.2 & 42.8 & \textit{71.5} & \textbf{80.6} & 431.2 \\
				& ~~+VL-NMS$_{\text{Trans}}$ (GT) & \textbf{72.1} & \textbf{90.9} & \textbf{95.0} & \textbf{45.8} & 69.8 & 78.0 & \textbf{451.6} \\
				\bottomrule
			\end{tabular}
		} 
	\end{center}
	\label{tab:itm_flickr}
    \vspace{-7pt}
\end{table*}

\begin{table*}[t]
	\small
	\caption{Performance of VL-NMS on ITM task with different matching backbones on MS-COCO. The metrics are Recall@K (\%) and Recall@Sum(\%). The best and second best methods under each setting are marked in bold and italic fonts, respectively.}
    \vspace{-7pt}
	\begin{center}
		\scalebox{0.95}{
			\begin{tabular}{l l  c c c  c c c  c}
				\toprule
				& \multirow{2}{*}{Models} & \multicolumn{3}{c}{Sentence Retrieval} & \multicolumn{3}{c}{Image Retrieval} & \multirow{2}{*}{R@sum} \\
				& & R@1 & R@5 & R@10 & R@1 & R@5 & R@10 \\
				\midrule
                \multicolumn{9}{c}{1K Test Images} \\
                \midrule
				\parbox[t]{2mm}{\multirow{8}{*}{\rotatebox[origin=c]{90}{N=10}}} & \cc{SCAN~\cite{lee2018stacked}} & \textit{\cc{47.5}} & \textit{\cc{77.3}} & \textit{\cc{87.0}} & \cc{34.7} & \cc{69.7} & \cc{81.4} & \cc{397.6} \\
				& \cc{~~+VL-NMS$_\text{FC}$ (GT)} & \cc{46.2} & \cc{75.6} & \cc{85.9} & \textit{\cc{37.3}} & \textit{\cc{72.4}} & \cc{83.3} & \textit{\cc{400.7}} \\
                & \cc{~~+VL-NMS$_\text{FC}$ (Transfer)} & \cc{45.7} & \cc{76.5} & \cc{86.4} & \cc{35.7} & \cc{71.2} & \textit{\cc{83.4}} & \cc{398.8} \\
				& \cc{~~+VL-NMS$_{\text{Trans}}$ (GT)} & \textbf{\cc{49.8}} & \textbf{\cc{80.2}} & \textbf{\cc{88.5}} & \textbf{\cc{39.5}} & \textbf{\cc{73.8}} & \textbf{\cc{85.6}} & \textbf{\cc{417.4}} \\
				\cmidrule(lr){2-9}
				& \cc{IMRAM~\cite{chen2020imram}} & \cc{41.9} & \cc{70.9} & \cc{80.9} & \cc{18.3} & \cc{50.3} & \cc{63.3} & \cc{325.6} \\
				& \cc{~~+VL-NMS$_\text{FC}$ (GT)} & \textit{\cc{46.9}} & \textit{\cc{75.4}} & \textit{\cc{86.5}} & \textbf{\cc{32.2}} & \textit{\cc{63.9}} & \textbf{\cc{76.2}} & \textit{\cc{381.2}} \\
                & \cc{~~+VL-NMS$_\text{FC}$ (Transfer)} & \cc{44.7} & \cc{76.7} & \cc{86.3} & \cc{30.9} & \cc{62.2} & \cc{75.1} & \cc{375.9} \\
				& \cc{~~+VL-NMS$_{\text{Trans}}$ (GT)} & \textbf{\cc{49.8}} & \textbf{\cc{78.1}} & \textbf{\cc{87.3}} & \textit{\cc{31.7}} & \textbf{\cc{64.0}} & \textbf{\cc{76.2}} & \textbf{\cc{387.2}} \\
                \midrule
                \multicolumn{9}{c}{5K Test Images} \\
                \midrule
                \parbox[t]{2mm}{\multirow{8}{*}{\rotatebox[origin=c]{90}{N=10}}} & \cc{SCAN~\cite{lee2018stacked}} & \textit{\cc{24.8}} & \textit{\cc{49.9}} & \textit{\cc{62.0}} & \cc{16.5} & \cc{40.0} & \cc{53.3} & \cc{246.4} \\
				& \cc{~~+VL-NMS$_\text{FC}$ (GT)} & \cc{24.2} & \cc{48.6} & \cc{61.2} & \textit{\cc{18.3}} & \textit{\cc{43.1}} & \textit{\cc{56.1}} & \textit{\cc{251.5}} \\
                & \cc{~~+VL-NMS$_\text{FC}$ (Transfer)} & \cc{22.5} & \cc{47.6} & \cc{61.2} & \cc{17.4} & \cc{41.5} & \cc{54.8} & \cc{245.1} \\
				& \cc{~~+VL-NMS$_{\text{Trans}}$ (GT)} & \textbf{\cc{28.2}} & \textbf{\cc{53.9}} & \textbf{\cc{65.8}} & \textbf{\cc{19.6}} & \textbf{\cc{44.9}} & \textbf{\cc{58.2}} & \textbf{\cc{270.6}} \\
				\cmidrule(lr){2-9}
				& \cc{IMRAM~\cite{chen2020imram}} & \cc{21.4} & \cc{45.1} & \cc{57.1} & \cc{10.1} & \cc{25.5} & \cc{34.7} & \cc{193.9} \\
				& \cc{~~+VL-NMS$_\text{FC}$ (GT)} & \textit{\cc{23.4}} & \textit{\cc{49.6}} & \textit{\cc{62.2}} & \textit{\cc{14.9}} & \textit{\cc{36.0}} & \textit{\cc{47.9}} & \textit{\cc{234.0}} \\
                & \cc{~~+VL-NMS$_\text{FC}$ (Transfer)} & \cc{21.7} & \cc{47.5} & \cc{60.8} & \cc{13.6} & \cc{33.6} & \cc{45.3} & \cc{222.4} \\
				& \cc{~~+VL-NMS$_{\text{Trans}}$ (GT)} & \textbf{\cc{26.2}} & \textbf{\cc{52.6}} & \textbf{\cc{64.9}} & \textbf{\cc{15.0}} & \textbf{\cc{36.7}} & \textbf{\cc{48.9}} & \textbf{\cc{244.4}} \\
				\bottomrule
			\end{tabular}
		} 
	\end{center}
	\label{tab:itm_coco}
    \vspace{-7pt}
\end{table*}

\subsection{Results on Image-Text Matching}

\textbf{Datasets.} \cc{We evaluated VL-NMS on two widely used ITM benchmark: Flickr30K~\cite{young2014image} and MS-COCO~\cite{lin2014microsoft}. Flickr30K consists of 31,000 images collected from Flickr, each associated with five human-annotated captions. Following the split in~\cite{lee2018stacked}, we used 1,000 images for validation, 1,000 images for test, and the rest for training. MS-COCO contains 123,287 images, and each image contains five captions. We follow the split in~\cite{lee2018stacked} and use 113,287 images for training, 5,000 images for validation and another 5,000 images for testing.}

\textbf{Evaluation Metrics.} Following~\cite{lee2018stacked}, we adopted Recall@K (R@K) to measure the performance of bidirectional retrieval, namely retrieving sentence with image query (sentence retrieval) and retrieving image with sentence query (image retrieval). R@K is defined as the fraction of queries for which the correct item is retrieved in the top-k scoring items. Also, to evaluate the overall retrieval performance, we report the R@Sum metric which is the sum of all R@K metrics as in~\cite{chen2020imram}.

\textbf{Settings.} \cc{To train VL-NMS, we need bbox annotations for all critical objects mentioned by the text query. However, ITM datasets don't provide such instance-level annotations. To address this issue, we explored two alternatives: \textit{1) By using extra annotations (VL-NMS (GT)):} For Flickr30K, we borrowed bbox annotations from Flickr30K Entities, which is a phrase grounding dataset built upon Flickr30K, as golden ground-truth for critical objects. For MS-COCO, we used the annotations for object detection to find caption-related objects. This can serve as the performance upper bound for VL-NMS on ITM task. \textit{2) By means of transfer learning (VL-NMS (Transfer)):} Given image-sentence pairs from the ITM dataset, we directly used a VL-NMS model trained on the REG dataset to generate proposals at the first stage. Specifically, we chose RefCOCOg as the source dataset as it contains longer expressions which is closer to the captions in Flickr30K and MS-COCO. We made this attempt in the belief that the fine-grained correspondence between image regions and text words learned by VL-NMS is fundamental and hence transferable. We experimented with using 10 proposals at the second stage, for reducing computation time and memory usage during inference.}

\textbf{Baselines.} We incorporated VL-NMS into two state-of-the-art two-stage ITM methods: \textbf{SCAN}~\cite{lee2018stacked} and \textbf{IMRAM}~\cite{chen2020imram}. Specifically, we used the SCAN~i-t~AVG model with the default configuration and Full-IMRAM with a single matching step due to memory constraints. 

\textbf{Results.} \cc{As is shown in Table~\ref{tab:itm_flickr} and  Table~\ref{tab:itm_coco}, all three methods can consistently improve the matching performance. VL-NMS$_{\text{Trans}}$ (GT) generally has higher performance improvements. We attribute it to the annotations of critical objects and performing the regression during training. It can be seen that reducing the number of proposals has a greater impact on the MS-COCO, since it generally has more caption-related objects (5 targets per sentence on average, while Flickr30K is 4.5 in our experiment). For SCAN, by integrating VL-NMS, using fewer proposals (\eg, 10 proposals) can achieve a close and even better performance than the original SCAN (36 proposals per image) on Flickr30K. For IMRAM, the performance drops drastically when reducing the number of proposals to 10. The results are worse than SCAN, suggesting that complex ITM methods are more sensitive to inferior proposals. Applying all three versions of VL-NMS can help capture meaningful visual context with a limited number of proposals, resulting in a huge performance boost over baseline (+59.1\% for VL-NMS$_\text{FC}$ (GT) and +45.6\% for VL-NMS$_\text{FC}$ (Transfer) and +66.0\% for VL-NMS$_{\text{Trans}}$ (GT) on R@sum metric on Flickr30K, +55.6\% for VL-NMS$_\text{FC}$ (GT) and +50.3\% for VL-NMS$_\text{FC}$ (Transfer) and +61.6\% for VL-NMS$_{\text{Trans}}$ (GT) on R@sum metric on MS-COCO).}

\begin{figure*}[!ht]
	\centering
	\includegraphics[width=1.0\linewidth]{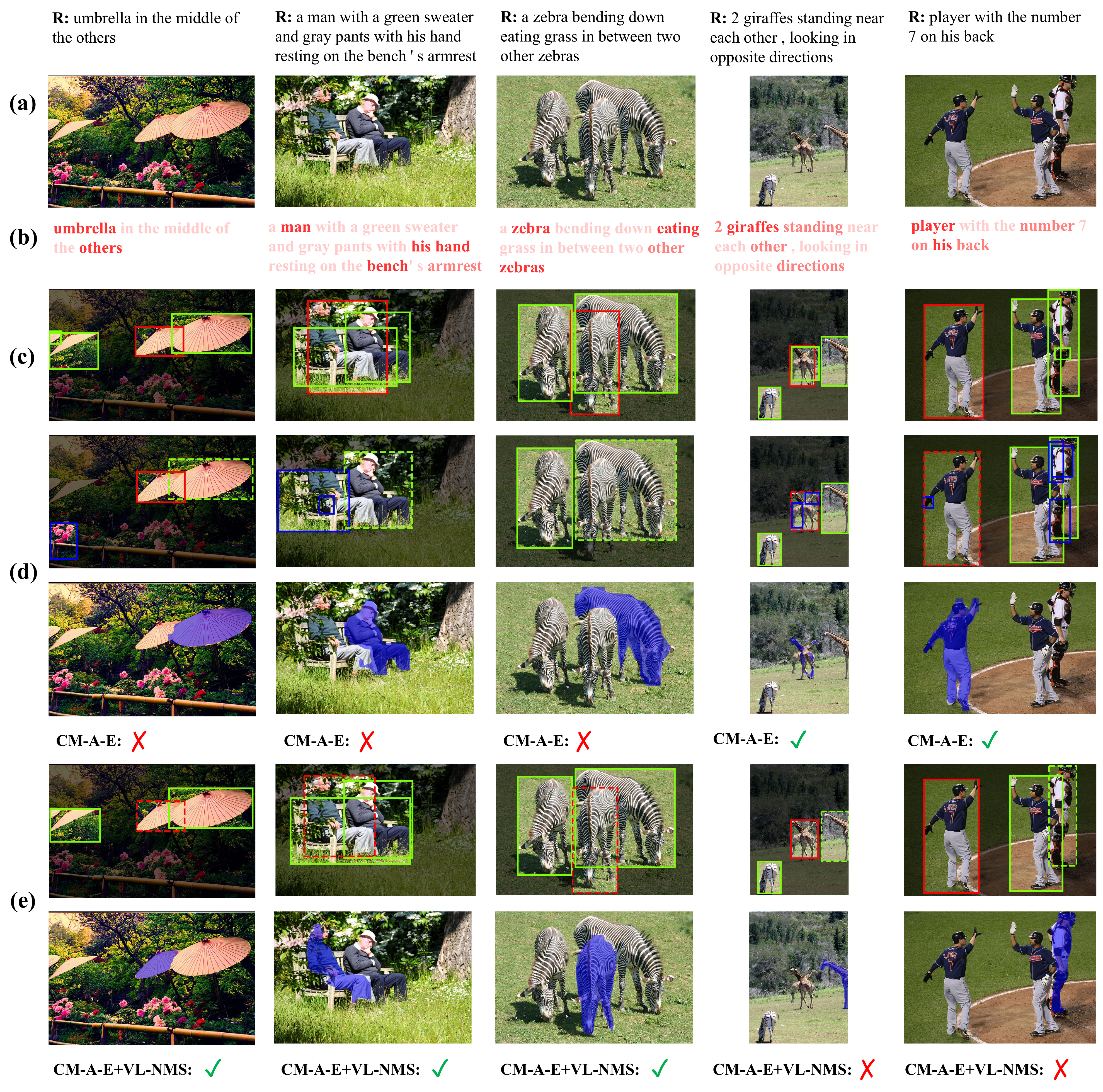}
	\caption{Qualitative REC and RES results on RefCOCOg showing comparisons between correct (green tick) and wrong referent grounds (red cross) by CM-A-E and CM-A-E+VL-NMS. \textbf{(a):} The referring expression and the input image. \textbf{(b):} The visualization of word attention weights $\alpha$ (\cf, Eq.~\eqref{eq:1}) for each referent object.\textbf{(c):} The annotated ground-truth bbox of the referent (marked in red) and the generated pseudo ground-truth bboxes of the contextual objects (marked in green). \textbf{(d):} The upper row demonstrates the proposals generated by an off-the-shelf object detector and the REC predictions of the downstream CM-A-E\cite{liu2019improving}; the lower row demonstrates the two-stage RES predictions acquired using the REC predictions from the upper row, the detailed method of which is fully described in \cite{yu2018mattnet}. \textbf{(e):} VL-NMS proposals, the REC and RES predictions from the downstream CM-A-E, arranged in the same format as \textbf{(d)}. The predicted bbox in REC is shown in dashed line. The denotations of bbox colors are as follows. \textbf{Red:} The bbox hits (IoU$>$0.5) the ground-truth bbox of the referent; \textbf{Green:} The bbox hits one of the pseudo ground-truth bboxes of the contextual objects; \textbf{Blue:} The false positive proposals.}
	\label{fig:visualization}
 \vspace{-15pt}
\end{figure*}

\subsection{Qualitative Results}

We illustrate the qualitative results between CM-A-E+VL-NMS and baseline CM-A-E on REC and RES tasks in Figure~\ref{fig:visualization}. From the results in line (b), we can observe that VL-NMS can assign high attention weights to words that are more relevant to individual referents (\eg, umbrella, man, and zebra). The results in line (c) show that the generated pseudo ground-truth bboxes can almost contain all contextual objects in the expression, except a few objects whose categories are far different from the categories of COCO (\eg, sweater, armrest, and grass). By comparing the results between line (d) and line (f), we have the following observations: 1) The baseline method tends to detect more false-positive proposals (\ie, the blue bboxes), and misses some critical objects (\ie, the red and green bboxes). Instead, VL-NMS can generate expression-aware proposals which are more relevant to the expression. 2) Even for the failure cases of CM-A-E+VL-NMS (\ie, the last two columns), VL-NMS still generates more reasonable proposals (\eg, with fewer false positive proposals), and the grounding errors mainly come from the second stage grounding model. The segmentation results of the two methods are shown in line (e) and line (g). We can observe that when extending REC models to the RES task, the quality of the segmentation mask is heavily dependent on the precision of REC predictions. If the REC model can ground the referent precisely with a bbox, the downstream segmentation branch can segment the referent from the bbox almost perfectly. Thus, by lifting the grounding precision of two-stage REC methods, VL-NMS can also help to improve the performance of two-stage RES methods.

\section{Conclusions and Future Works}
In this paper, we focused on the two-stage visual-language grounding and matching, and discussed the overlooked mismatch problem between the roles of proposals in different stages. Particularly, we proposed a novel algorithm VL-NMS to calibrate this mismatch. VL-NMS tackles the problem by considering the query at the first stage, and learns a relatedness score between each detected proposal and the query. The product of the relatedness score and classification score serves as the suppression criterion for the NMS operation. Meanwhile, VL-NMS is agnostic to the downstream grounding and matching step, hence can be integrated into any SOTA two-stage grounding and matching method. Extensive ablations on various tasks and benchmarks consistently demonstrate that VL-NMS is robust, generalizable, and transferable. Moving forward, we plan to apply VL-NMS into other proposal-drive visual-language tasks which suffer from the same mismatch issue, \eg, video grounding~\cite{xiao2021boundary,chen2020rethinking,cao2021pursuit,xiao2021natural}, image captioning~\cite{yu2019multimodal,mao2022rethinking} and VQA~\cite{chen2020counterfactual,chen2021counterfactual,yu2019multi,li2019semantic, yu2018beyond, yu2022knowledge, chen2022rethinking}.


\begin{acks}
This work was supported by the National Key Research \& Development Project of China (2021ZD0110700), the National Natural Science Foundation of China (U19B2043, 61976185), Zhejiang Innovation Foundation (2019R52002), and the Fundamental Research Funds for the Central Universities (226-2022-00051).
\end{acks}

\bibliographystyle{ACM-Reference-Format}
\bibliography{main.bib}

\end{document}